\documentclass{article}
\usepackage[T1]{fontenc}
\usepackage{iclr2027_conference,times}

\usepackage{amsmath,amsfonts,bm}

\def\eqref#1{equation~\ref{#1}}

\def\1{\bm{1}}

\DeclareMathAlphabet{\mathsfit}{\encodingdefault}{\sfdefault}{m}{sl}
\SetMathAlphabet{\mathsfit}{bold}{\encodingdefault}{\sfdefault}{bx}{n}

\usepackage{enumitem}
\usepackage{graphicx}
\usepackage{listings}
\lstdefinestyle{prompt}{basicstyle=\footnotesize\ttfamily,breaklines=true,
  columns=fullflexible,keepspaces=true,showstringspaces=false,
  literate={count,}{{\mbox{count,}}}6,
  aboveskip=6pt,belowskip=6pt}
\usepackage{booktabs}
\usepackage{xcolor}
\usepackage{hyperref}
\usepackage{url}
\definecolor{LabBlue}{HTML}{3B6694}
\definecolor{LabPurple}{HTML}{7D527B}
\definecolor{AwardOrange}{HTML}{A55D0F}
\definecolor{ForestGreen}{HTML}{2F6B4F}
\definecolor{LabDark}{HTML}{303744}
\hypersetup{
  colorlinks=true,
  linkcolor=LabBlue,
  citecolor=LabPurple,
  urlcolor=AwardOrange,
  filecolor=ForestGreen,
  anchorcolor=LabDark,
  breaklinks=true
}
\title{When Can Text Replace Vision? Structural Bottlenecks in Diagram Reasoning}
\author{Yunbei Zhang$^{1}$\thanks{Corresponding author: \texttt{yzhang111@tulane.edu}.} \quad Janet Wang$^{1}$ \quad Jihun Hamm$^{1}$ \quad Chandan K. Reddy$^{2}$\\
\textnormal{$^{1}$Tulane University} \quad
\textnormal{$^{2}$Virginia Tech}\\[0.4cm]
\textnormal{ Code: \url{https://github.com/yunbeizhang/text-for-vision}}}
\hypersetup{pdfauthor={Yunbei Zhang, Janet Wang, Jihun Hamm, Chandan K. Reddy}}
\iclrfinalcopy
\begin{document}
\maketitle
\lhead{Preprint}

\begin{abstract}
Can structured text replace vision for diagram reasoning? A wrong answer after
textualization can arise because the representation omits information the question
needs, or because the solver fails to use information that is present. We introduce
a diagnostic protocol to distinguish these explanations. Using the same solver
model and generation settings, we compare three input conditions: the original
image, question-blind structure extracted by
a vision-language model, or gold structure derived from the diagram source.
Validity-triggered recovery tests truncation and schema failure, question-relevant
fidelity measures preservation of answer-critical structure, and matched edge
interventions test the effect of error location. On a reserved holdout of 240 public
FlowGen diagrams, evaluated under a frozen protocol, gold structure reaches 87\%
accuracy while direct vision and learned text both remain below 30\%. The
aggregate comparison includes source-derived relation labels that may not be
printed in the image and uses different learned and gold graph encodings,
so it does not isolate extraction error alone. Retrying
only invalid extractions makes nearly every public representation schema-valid
yet leaves accuracy essentially unchanged; in a post-confirmation diagnostic on
generated diagrams, the same policy brings learned-text accuracy close to
direct vision. The public learned-text deficit relative to gold more than doubles with
structural difficulty. Question-relevant topology predicts correctness better than
whole-graph topology. In an exposed intervention study, a single answer-relevant
edge edit reduces the primary solver's original-answer accuracy to near zero, while matched
irrelevant edits largely preserve it. Supplied structure requires fewer solving
tokens than vision, but learned acquisition removes this advantage at single
use. These comparisons motivate evaluating acquired text by the answer-relevant
evidence it preserves and by the solver's ability to use that representation.
\end{abstract}
\section{Introduction}

Can structured text replace vision for diagram reasoning?
A diagram often encodes the answer to a question in a small set of relations. If those relations can be recovered as text, a language model can reason over an explicit representation that can be reused across questions and may require fewer solving tokens than the original image. This idea underlies structured visual interfaces such as plot-to-table conversion in DePlot \citep{liu2023deplot}, screenshot parsing in Pix2Struct \citep{lee2023pix2struct}, and language-mediated composition of perception and reasoning models \citep{zeng2022socraticmodels}. Chart-derendering pretraining also strengthens visual reasoning \citep{liu2022matcha}.
\begin{figure}[t]
\centering
\includegraphics[width=\linewidth]{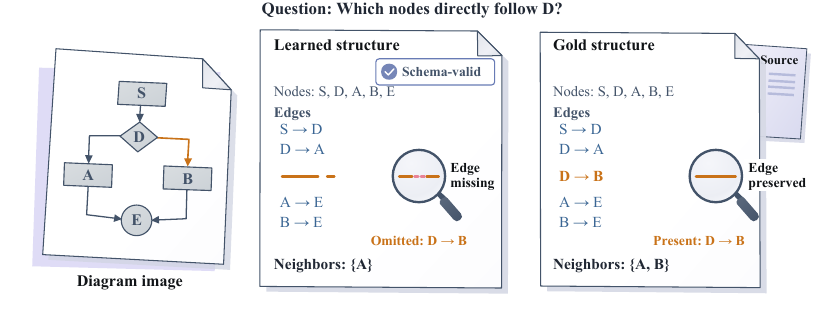}
\caption{\textbf{Schema-valid text can still omit answer-critical structure.} For the question shown, the diagram and source give $\{A,B\}$. A learned representation can retain every node yet omit $D\!\to\!B$, yielding only $\{A\}$. The highlighted relation determines whether the textual representation preserves the answer. The example motivates evaluating text substitution by the structure needed for the question, rather than node coverage or parse validity alone. This is an illustrative example; edge lists are schematic rather than recorded model outputs.}
\label{fig:teaser}
\end{figure}

The difficulty is that producing structured text is not the same as producing the \emph{right} structured text. A transcription can be syntactically valid and globally similar to the source while omitting the one relation needed to answer a question. Conversely, substantial errors elsewhere in the graph may not affect that answer at all. End-task accuracy alone does not reveal whether
text failed because the necessary information was never recovered or because the solver could not use information that was present; whole-graph fidelity does not reveal whether the missing information mattered to the question. MathVista documents joint perception and reasoning challenges \citep{lumathvista}, while MathVerse varies the information supplied through text and diagrams to test visual dependence \citep{zhang2024mathverse}. We therefore distinguish \emph{representation sufficiency} from \emph{representation acquisition}, and test how preserving answer-relevant structure relates to reasoning success.

We study this distinction using a fixed-solver protocol with three matched input conditions: the original image, question-blind structured text (extracted without access to the question), and gold structure derived from the diagram source. The gold condition tests the utility of supplied textual structure for the question; the learned condition measures how much utility the image-to-text pipeline achieves. On a reserved holdout of 240 public FlowGen diagrams \citep{shi2026flowgen}, the distinction is stark: gold structure reaches 87\% accuracy, while direct vision and learned text both remain below 30\%. This establishes a large utility difference between supplied source structure and the tested image-based routes, without by itself isolating the cause.

The public acquisition gap survives bounded, validity-triggered recovery. Retrying only invalid extractions makes nearly all public representations valid with little change in QA; already-valid extractions are unchanged. In contrast, the same policy brings learned-text accuracy close to direct vision in a post-confirmation diagnostic on generated diagrams. The public comparison includes source-derived relation labels that may not be printed in the image and different graph encodings; Sec.~\ref{sec:gap} explains their implications for the measured accuracy deficit.

The learned-text deficit relative to gold more than doubles across node-count bins and grows with branch depth. Question-relevant topology predicts QA better than whole-graph topology on matched holdout cases. Exposed interventions provide complementary mechanistic evidence: for the primary 122B solver, changing one answer-relevant edge nearly eliminates original-answer accuracy, whereas matched irrelevant edits largely preserve it. What matters is not simply how much structure is recovered, but whether it contains what the question requires. As a secondary systems result, supplied structure requires fewer solving tokens than images, but learned acquisition removes this advantage at single use. Reuse lowers cost without closing the public accuracy deficit. Our contributions are:
\begin{enumerate}[label=(\arabic*),leftmargin=2.2em,itemsep=0.4em]
    \item \textbf{A diagnostic protocol separating sufficiency from acquisition.} A fixed-solver input ladder and validity recovery distinguish the utility of supplied structure from failures in acquiring it.

    \item \textbf{A difficulty-dependent gap confirmed on a reserved holdout.} Public learned--gold gaps survive recovery and widen with node count and branch depth, unlike controlled recovery.

    \item \textbf{Predictive and intervention evidence for answer-relevant structure.} Relevant topology predicts QA better than whole topology, and matched interventions show that error location matters beyond edit count.
\end{enumerate}
\section{A Protocol for Separating Sufficiency from Acquisition}
\label{sec:protocol}

\subsection{Fixed-solver input ladder}
\label{sec:protocol_ladder}

Following image-to-text pipelines such as DePlot \citep{liu2023deplot},
our fixed-solver ladder adds a source-structure reference to evaluate acquired text.

Let $I_g$ denote a diagram image, $G_g$ its released source structure, and
$(q,y)$ a question and reference answer. A question-blind extractor $E_b$
produces $z_g=E_b(I_g)$ without access to $q$ or $y$, where $b$ specifies the
acquisition policy and its output-token budget. Holding the downstream solver
$S$ fixed, we compare three matched input conditions:
\[
\hat y_{\mathrm{direct}} = S(I_g,q), \qquad
\hat y_{\mathrm{learned}} = S(z_g,q), \qquad
\hat y_{\mathrm{gold}} = S(G_g,q).
\]
Learned and gold conditions receive serialized structure without the image;
the solver must still derive the answer, with its procedure and output cap
fixed across conditions. Both share an outer JSON contract and compact serialization, but differ in
graph encoding: learned inputs use
entity and relation objects, whereas public gold inputs embed labeled source
triplets in a JSON string. Thus, the acquisition gap compares utility under
these recorded input constructions, not extraction error under a controlled
common graph encoding. Exact prompts and inputs appear in Appendix~\ref{app:input-contracts}.

Gold tests \emph{representation sufficiency} for the solver; learned text
additionally requires \emph{acquisition} from pixels. For normalized
exact-match accuracy $A_a$ under condition $a$, define
\[
    \Delta_{\mathrm{acq}}
    = A_{\mathrm{learned}} - A_{\mathrm{gold}}, \qquad
    \Delta_{\mathrm{suff}}
    = A_{\mathrm{gold}} - A_{\mathrm{direct}},
\]
the acquisition and sufficiency gaps, respectively. A more negative
$\Delta_{\mathrm{acq}}$ indicates a larger utility deficit of the learned
pipeline relative to the source-derived reference;
a positive $\Delta_{\mathrm{suff}}$ means source structure improves on direct
vision. This empirical sufficiency gap depends on the solver,
serialization, cap, and questions; it is not universal.

\subsection{Validity recovery and structure-sensitive evaluation}
\label{sec:protocol_recovery}
\label{sec:protocol_fidelity}
\label{sec:protocol_interventions}

\textbf{Validity recovery.}
To test whether truncation or malformed JSON explains the acquisition gap,
we retry only schema-invalid extractions using a fixed sequence of larger
budgets, then an answer-free normalizer and at most one final retry.
The policy never uses the question, answer, or downstream QA to select text;
valid but semantically incorrect extractions are retained. It tests bounded
validity recovery, not uniform re-extraction at the largest budget.
Caps are specified in Sec.~\ref{sec:setup}.

\textbf{Question-relevant fidelity.}
For each question, an offline support mask $M(q,G_g)$ identifies incident
edges for neighbor queries, a labeled ordered pair for relation queries, or
a conservative search certificate for shortest-path queries. Masks are
withheld from extractor and solver and need not be minimal sufficient
subgraphs. We compare relevant and whole directed topology exact match
(primary) and edge F1 (secondary) on identical learned representations.
Directed topology exact requires equality of normalized node sets and
directed-edge sets within the evaluated scope, ignoring edge labels;
label-sensitive fidelity is evaluated separately. Matched one-feature predictors share chart-held-out folds; lower out-of-fold
Brier loss indicates better QA prediction. AUC is secondary.

\textbf{Evaluation populations.}
End-to-end QA and fidelity prediction use different populations because they
answer different questions. QA includes all intended questions and treats an
invalid acquisition as a failure, measuring whether a policy delivers a
usable answer. The matched-valid fidelity comparison instead asks which
property of an acquired representation predicts correctness once the input
can be evaluated. Using identical cases for relevant and whole metrics
prevents differences in validity coverage from driving their comparison.
A stronger predictor on this subset is evidence about representation
utility, not a substitute for the all-intended accuracy of the pipeline.

\textbf{Matched interventions.}
To complement predictive association, we corrupt source structure in matched
pairs: one or two deletions, reversals, or redirects. Relevant edits change
the original answer semantics; matched irrelevant edits preserve them.
Both conditions are scored against the original answer. Matching type and
count isolates the placement of corruption, not its nominal size. This
exposed mechanism study measures sensitivity to deliberately answer-changing
edits, not their contribution to the natural acquisition gap. Recovery,
support-mask construction, and intervention rules are documented in
Appendices~\ref{app:recovery}, \ref{app:fidelity}, and~\ref{app:mechanism}.

\subsection{Confirmatory criteria and cost}
\label{sec:protocol_criteria}

The three prespecified primary tests concern a more negative acquisition
gap with node count and branch depth, and better QA prediction from relevant
than whole topology. Recovery and the 27B solver arm are secondary;
question-family analyses are exploratory. Main gap interpretations require
gold QA of at least 70\% in a prespecified cell. Text is comparable to vision
only if the lower paired 95\% confidence bound for
$A_{\mathrm{text}}-A_{\mathrm{direct}}$ exceeds $-2$ percentage points;
non-significance alone does not establish comparability.

We account for the full acquisition cost of learned representations, including
failed attempts. Let $A_g^{\mathrm{tok}}$ denote all acquisition prompt and
output tokens for diagram $g$, and let $S_{gq}^{\mathrm{tok}}$ denote solver
tokens for question $q$. We report single-question use ($K=1$) and reuse
across the three observed questions ($K=3$), with average cost
\[
    C_K(g) =
    \frac{
        A_g^{\mathrm{tok}}
        + \sum_{q=1}^{K} S_{gq}^{\mathrm{tok}}
    }{K}.
\]
Gold is a solving-only reference with zero modeled acquisition cost because its structure is supplied.
\section{Experimental Setup}
\label{sec:setup}

\textbf{Public data and tasks.}
Our confirmatory evaluation uses FlowGen's released flowchart images and source graphs \citep{shi2026flowgen}. We reserve 240 previously unevaluated charts from its official \emph{Diagrams} renderer, with 80 from each released easy, medium, and hard stratum, using the images unchanged. Reservation excludes exact and near-duplicate groups containing exposed charts, without using model outcomes. Three deterministic source-derived questions per chart yield 720 questions across four families: incoming neighbors, outgoing neighbors, source edge relations, and shortest-path length. These ask for a node's predecessor or successor set, the source relation on an ordered node pair, or the number of edges in their shortest directed path; they are not FlowGen's native questions. Answers are checked against source structure, which can include relations not explicitly labeled in the image. They test both local relation recovery and multi-edge traversal, rather than treating every diagram question as the same reasoning task. A separate exposed cohort of 240 charts and 720 questions supports recovery diagnostics. Interventions use a preselected 60-chart subset, with eligibility varying by edit type and count. Dataset examples and reservation details appear in Appendices~\ref{app:datasets} and~\ref{app:protocol}.

\textbf{Models and acquisition policies.}
The primary extractor and solver are Qwen3.5-122B-A10B, a mixture-of-experts model with approximately 10B active parameters \citep{qwen2026model}. The question-blind extractor uses a fixed generic JSON-transcription instruction and a 1,024-token output cap. Secondary recovery accepts the first valid representation, escalating invalid outputs to 2,048 and then 8,192 tokens, followed if needed by normalization and one image-and-validation-feedback retry. All confirmatory solving conditions use an 8,192-token cap, BF16 inference, temperature zero, seed 427, and thinking disabled. This common solving cap was selected using exposed cap checks before holdout evaluation; it is distinct from the budget for acquiring the representation. Extraction is performed without the downstream question, and the same resulting text can be supplied to each question about that chart. A secondary Qwen3.5-27B solver receives exactly the same fixed-extraction strings as 122B, not recovered representations. This comparison changes the solver while holding acquired information constant. Table~\ref{tab:modelsettings} lists settings for all comparisons.

\textbf{Controlled and secondary comparisons.}
Two controlled generated cohorts contrast with public graph queries. The first contains 136 executable program flowcharts and 135 Boolean circuits; the second adds 270 program flowcharts with 20--40 nodes and branch depth 4--6. Flowchart questions ask for a designated variable's final value given initial integers; circuit questions ask for the output bit given printed inputs and gates. Source execution supplies the reference answers. Unlike the public neighbor and relation queries, these tasks require evaluating a program or circuit represented by the diagram. They therefore provide a contrasting setting in which to ask whether repairing representation validity recovers downstream utility. The original controlled confirmation was mixed and ceiling-limited; subsequent recovery is a post-confirmation diagnostic, not additional difficulty confirmation. These results are not pooled with public FlowGen. Exposed interventions use Qwen3.5-4B, 27B, and 122B-A10B at a common 512-token solver cap. The secondary QZhou comparison \citep{kingsoftqzhou} uses 200 charts and 600 native questions at the same cap.

\textbf{Measures and statistical analysis.}
QA uses normalized exact match; paired learned--gold differences are the primary difficulty outcome. Invalid acquisitions count as answer failures. Missing responses enter full-sample bounds, while paired intervals use complete observations. Node count and branch depth define the analysis bins, distinct from FlowGen's released strata; depth is the maximum number of branching vertices along a path after collapsing strongly connected components. Fidelity prediction uses matched schema/adapter-valid examples in five chart-held-out folds. Chart-cluster bootstrap intervals use 10,000 resamples and seed 427, keeping each chart's questions and conditions together: Bonferroni-adjusted 98.33\% intervals for the three primary tests and 95\% otherwise. Hypotheses, policies, bins, gold floor, comparability margin, and analysis were frozen before holdout access. Set-F1 and Jaccard rescoring are post-hoc sensitivities. The appendix records scoring, missingness, and inference details.

\section{Results}
\label{sec:results}

\subsection{A large public acquisition gap survives validity recovery}
\label{sec:gap}

Supplied structure supports high accuracy, but learned acquisition recovers little of that utility on the public FlowGen holdout. In Table~\ref{tab:main}, gold achieves 87.1\% QA with the same solver, versus 22.8\% for fixed learned text and 27.2--28.8\% for direct vision. The paired acquisition gap is $\Delta_{\mathrm{acq}}=-64.31$ percentage points (95\% CI $[-68.89,-59.58]$); the paired sufficiency gap is $\Delta_{\mathrm{suff}}=+59.24$ points. Paired comparisons also establish a smaller direct-over-learned advantage, with intervals in Fig.~\ref{fig:paired}. The direct range reflects bounds on 11 missing responses, not sampling uncertainty. Because learned and gold inputs use different graph encodings, the measured gap reflects both the tested acquisition pipeline and representation--solver compatibility, rather than extraction error alone.

A post-hoc audit found that relation targets need not be printed in the image: 124 of 205 relation questions target \texttt{connectedTo}, and 18 target \texttt{partOf}. Image spot checks show unlabeled links and container membership, respectively. On the \texttt{connectedTo} questions, gold is correct on 123/124, versus none for direct or either 122B learned policy. The frozen results thus include access to source-specific conventions, not just image-grounded extraction. Appendix~\ref{app:visibility-audit} documents the target-label audit and literal-label scoring. A visibility-qualified analysis would be required to isolate the contribution of source-specific conventions; we therefore do not interpret the aggregate gap as a pure image-extraction effect. The deficit is not confined to source edge relation questions: under frozen exact-match scoring, gold and fixed learned QA are 86.0\% and 25.3\% on incoming-neighbor questions, and 94.6\% and 23.5\% on outgoing-neighbor questions, respectively.

\begin{table}[!ht]
\caption{\textbf{Supplied structure substantially outperforms learned acquisition.} Public holdout: 240 charts, 720 questions per arm. Invalid acquisitions count as QA failures; direct bounds cover 11 missing responses. The secondary 27B solver receives the same fixed extraction.}
\label{tab:main}
\begin{center}
\small
\begin{tabular}{@{}lrrrr@{}}
\toprule
Input / solver & Valid / 240 & Correct & Missing & QA (\%) \\
\midrule
Direct vision & -- & 196 & 11 & 27.2--28.8 \\
Fixed learned & 196 & 164 & 0 & 22.8 \\
Recovered learned & 238 & 169 & 0 & 23.5 \\
Gold structure & -- & 627 & 0 & 87.1 \\
\midrule
Fixed learned $\rightarrow$ 27B & 196 & 157 & 0 & 21.8 \\
\bottomrule
\end{tabular}

\end{center}
\end{table}

\begin{figure}[t]
\centering
\includegraphics[width=\linewidth]{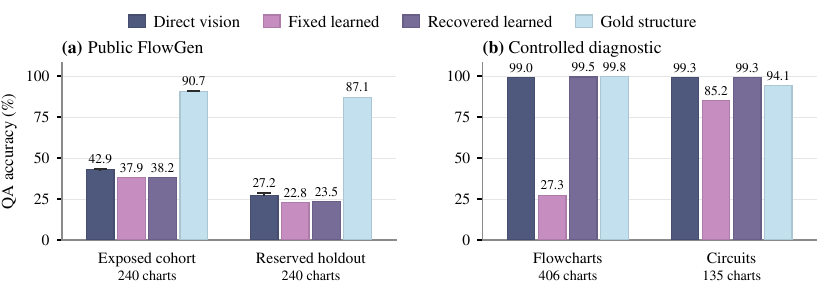}
\caption{\textbf{Recovery closes the controlled deficit, not the public gap.} (a) Separate exposed and held-out FlowGen cohorts, each 240 charts and 720 questions. Whiskers bound missing outcomes (holdout: 11 direct; exposed: 3 direct, 2 gold), not confidence intervals. (b) Post-confirmation recovery on controlled program flowcharts and Boolean circuits. The four conditions compare the original image, fixed learned text, recovered learned text, and source-derived gold. Recovery retries only invalid extractions; already-valid text is unchanged. Both panels use Qwen3.5-122B-A10B and an 8,192-token solver cap.}
\label{fig:recovery}
\end{figure}

Validity recovery barely improves public QA. It increases schema-valid representations from 196/240 to 238/240 charts, but accuracy rises by only 0.7 points. Fig.~\ref{fig:recovery} shows the same pattern on the separate exposed cohort: all charts become valid with little QA improvement. A successful parse establishes that a representation can be consumed, not that its nodes, edges, and labels reproduce the diagram. Recovery substantially reduces the first obstacle without removing the downstream deficit. This diagnostic holds under the tested source-derived targets and input encodings; it does not separate their contribution from semantic extraction errors. Because already-valid extractions are unchanged, it also does not rule out gains from uniform high-budget re-extraction or semantic refinement.

Partial-credit scoring preserves the gold--direct--learned ordering. Replacing strict neighbor-set equality with set-F1 or thresholded Jaccard raises scores without eliminating either the gold advantage or the smaller direct-over-learned advantage; paired intervals exclude zero under both variants. Relation and shortest-path scoring remain unchanged. The public deficit persists beyond exact neighbor-set scoring. Appendix~\ref{app:lenient} reports both sensitivities for all five arms.

Controlled recovery provides the contrasting diagnostic in the same figure. All 541 representations become valid, and learned QA reaches near-ceiling accuracy comparable to direct vision on program flowcharts and Boolean circuits. Truncation, schema, and acquisition failures therefore explain much of their low-budget deficit: an initial learned--gold gap does not by itself establish a persistent semantic limitation. This follows a mixed, ceiling-limited controlled confirmation and remains a post-confirmation diagnostic, not another confirmatory test or part of the public aggregate. Appendix~\ref{app:controlled} reports task-level results.

\subsection{Structural difficulty widens the acquisition gap}
\label{sec:difficulty}

\begin{figure}[t]
\centering
\includegraphics[width=\linewidth]{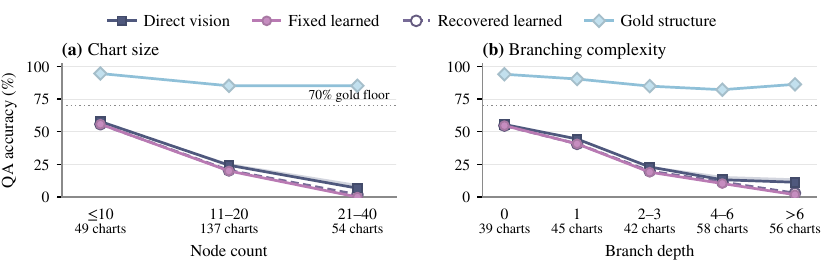}
\caption{\textbf{The public acquisition gap widens with chart size and branch depth.} QA on the 240-chart public holdout, grouped by prespecified node-count and branch-depth bins. All conditions use the same Qwen3.5-122B-A10B solver at an 8,192-token cap; direct shading denotes missing-outcome bounds. Fixed and recovered learned curves nearly overlap, while gold remains above the 70\% floor (dotted) in every occupied bin. Lines connect bins; they are not fitted curves.}
\label{fig:difficulty}
\end{figure}

The public acquisition gap widens sharply with chart size and branch depth. Fig.~\ref{fig:difficulty} shows the learned--gold deficit growing from 38.8 points on charts with at most ten nodes to 85.2 points on charts with 21--40 nodes. Recovery barely changes this pattern. Both prespecified difficulty trends confirm after multiplicity adjustment: learned utility declines relative to gold across ordered node-count and depth bins. Direct vision also deteriorates faster than gold, indicating that difficulty affects both routes starting from pixels while source structure remains comparatively usable. Comparing learned and gold answers on the same questions is essential here. A decline in learned QA alone could reflect harder downstream reasoning; the widening paired gap instead shows an increasing utility deficit relative to the supplied source representation.

These are equal-chart observational trends across bins, not effects per additional node or unit of depth; node count and depth can covary. All occupied bins pass the 70\% gold floor. However, exploratory shortest-path gold is 66.7\%, below that floor, leaving a solver-side residual under the tested representation and cap. Those questions remain in aggregate analyses but do not support an exclusively extraction-limited family-level interpretation. These trends also include the source-derived relation targets noted in Sec.~\ref{sec:gap}; their interpretation is conditional on that question construction. Per-bin gaps and adjusted trend intervals appear in Tables~\ref{tab:difficulty} and~\ref{tab:inference}.

\subsection{Question-relevant topology predicts downstream correctness}
\label{sec:fidelity}

Question-relevant topology exact match predicts QA better than whole-graph topology exact match. In Table~\ref{tab:fidelitymain}, restricting evaluation to question support reduces out-of-fold Brier loss from 0.1406 to 0.1147 on the same valid fixed-extraction cases; the multiplicity-adjusted interval excludes zero. The predictors use identical observations, chart-held-out folds, and one-feature logistic fitting. Only the metric's scope changes; source-derived masks are never supplied to extractor or solver. Unlike whole-graph exact match, relevant topology distinguishes an omitted unrelated branch from an omitted answer-critical edge. Its lower prediction loss shows that this distinction carries information about downstream success.

This advantage is metric-dependent: matched edge F1 does not establish an improvement. Exact preservation of the evaluated support and average overlap with that support are different criteria; the evidence for the former should not be generalized to every relevance-restricted metric. Nor does a topology match guarantee a correct answer, since labels and the solver's use of the representation also matter. Recovered inputs also show lower relevant-topology loss, a descriptive secondary result. Appendix~\ref{app:fidelity} reports matched sensitivities, AUC, and label-sensitive results.

\begin{table}[t]
\caption{\textbf{Relevant topology improves QA prediction.} Lower Brier loss is better; $\Delta$ is relevant minus whole on matched cases and folds. Intervals are 98.33\% for the primary test ($\dagger$), 95\% otherwise. Recovery and edge F1 are secondary; edge-F1 intervals cross zero.}
\label{tab:fidelitymain}
\begin{center}
\small
\setlength{\tabcolsep}{3.5pt}
\begin{tabular}{@{}llrrr@{\hspace{8pt}[}r@{, }r@{]}l@{}}
\toprule
Extraction & Fidelity metric & Whole & Relevant & \multicolumn{1}{c}{$\Delta$} & \multicolumn{2}{c}{Paired CI} & \\
\midrule
Fixed (588 Q) & Topology exact & 0.1406 & 0.1147 & $-$0.0259 & $-$0.0479 & $-$0.0047 & $^{\dagger}$ \\
 & Edge F1 & 0.1305 & 0.1240 & $-$0.0065 & $-$0.0162 & 0.0032 & \\
\addlinespace[3pt]
Recovered (714 Q) & Topology exact & 0.1242 & 0.1058 & $-$0.0184 & $-$0.0346 & $-$0.0029 & \\
 & Edge F1 & 0.1165 & 0.1134 & $-$0.0031 & $-$0.0115 & 0.0054 & \\
\bottomrule
\end{tabular}

\end{center}
\end{table}

\subsection{Matched interventions reveal the effect of error location}
\label{sec:mechanism}

Answer-critical corruption is much more damaging than answer-preserving corruption at the same edit type and count. In Table~\ref{tab:perturbationmain}, a single relevant edge edit yields 0--3.5\% QA for 122B, whereas matched irrelevant edits preserve 92.4--93.9\%. Reversal, deletion, and redirection show the same pattern with one or two edits. The direction holds across all three solvers, although 4B is less accurate even under irrelevant edits. Error count alone is therefore an inadequate description of downstream impact. Unlike the fidelity analysis, which observes naturally extracted representations, this comparison deliberately changes where corruption occurs while matching its nominal type and size. It provides intervention evidence that complements the predictive association: a small structural error can be consequential when it changes the evidence required by the question.

This is an exposed FlowGen mechanism study at a 512-token solver cap, separate from the 8,192-cap confirmation. Each matched pair starts from the same source structure. Relevant edits intentionally change the original answer, while irrelevant edits preserve it; both are scored against that original answer. The contrast measures sensitivity to selected structural corruption, not correctness on the modified graph or the fraction of the natural acquisition gap caused by such errors. Appendix~\ref{app:mechanism} gives paired intervals; Appendix~\ref{app:qualitative} illustrates recorded errors and preserved support.

\begin{table}[!ht]
\caption{\textbf{Relevant corruption is more damaging than matched irrelevant corruption.} Exposed FlowGen, 512-token solver cap; QA (\%) is scored against original answers. Each row matches chart/question pairs and edit count $m$. The 4B solver is lower even on irrelevant edits.}
\label{tab:perturbationmain}
\begin{center}
\small
\begin{tabular}{@{}lrrrrrrrrr@{}}
\toprule
 & & & & \multicolumn{2}{c}{4B} & \multicolumn{2}{c}{27B} & \multicolumn{2}{c}{122B} \\
\cmidrule(lr){5-6}\cmidrule(lr){7-8}\cmidrule(l){9-10}
Edit & $m$ & Charts & Q & Rel. & Irrel. & Rel. & Irrel. & Rel. & Irrel. \\
\midrule
Reverse & 1 & 60 & 172 & 5.2 & 78.5 & 4.7 & 93.6 & 3.5 & 92.4 \\
Reverse & 2 & 58 & 122 & 3.3 & 68.9 & 1.6 & 92.6 & 0.0 & 91.0 \\
\addlinespace[3pt]
Drop & 1 & 50 & 148 & 0.0 & 79.1 & 0.0 & 93.9 & 0.0 & 93.9 \\
Drop & 2 & 49 & 100 & 0.0 & 78.0 & 0.0 & 93.0 & 0.0 & 94.0 \\
\addlinespace[3pt]
Redirect & 1 & 50 & 148 & 0.7 & 74.3 & 0.0 & 92.6 & 0.0 & 93.2 \\
Redirect & 2 & 49 & 100 & 0.0 & 72.0 & 0.0 & 91.0 & 0.0 & 91.0 \\
\bottomrule
\end{tabular}

\end{center}
\end{table}

\subsection{Cost and solver compatibility}
\label{sec:secondary}

Acquisition removes the single-use token advantage. Table~\ref{tab:costmain} shows that supplied gold structure requires 81.8\% fewer solving tokens than direct vision, but learned acquisition, including failed attempts, makes both learned policies more token-intensive at $K=1$. Reuse across three questions lowers usage without repairing missing information; neither policy meets the predeclared comparable-accuracy margin. An efficient textual interface therefore need not yield efficient end-to-end substitution. These are native-token savings, not dollar or hardware-controlled compute savings, and gold excludes acquisition. Appendix~\ref{app:cost} provides the full cost accounting.

\begin{table}[t]
\caption{\textbf{Reuse saves tokens without comparable QA.} Means over 720 intended holdout questions include failed acquisitions. $K=3$ reuses the same representation and answers; gold excludes acquisition. Only acquisition is amortized; direct and gold retain their per-question solving costs. The lower paired 95\% text-minus-direct bound must exceed $-2$ points for comparability.}
\label{tab:costmain}
\begin{center}
\small
\setlength{\tabcolsep}{5pt}
\begin{tabular}{@{}lrrrc@{}}
\toprule
Input & QA (\%) & Tokens/Q, $K=1$ & Tokens/Q, $K=3$ & Comparable QA? \\
\midrule
Direct vision & 27.2--28.8 & 6,423 & 6,423 & -- \\
Fixed learned & 22.8 & 7,551 & 3,201 & No \\
Recovered learned & 23.5 & 10,413 & 4,397 & No \\
Gold structure & 87.1 & 1,169 & 1,169 & Yes \\
\bottomrule
\end{tabular}

\end{center}
\end{table}

On identical fixed extractions, the holdout 27B-minus-122B difference is $-0.97$ points (95\% CI $[-2.22,0.28]$). This does not establish a solver advantage; holding extraction bytes fixed tests representation--solver compatibility rather than comparing acquisition pipelines. Candidate solvers should therefore be evaluated on the actual extracted inputs.

QZhou offers an exposed counterpoint to FlowGen: at the secondary 512-token solver cap, direct vision scores 69.8\% versus 67.7\% with source structure; learned extraction at a 2,048-token cap reaches 71.0\%, a numerical difference without an established superiority claim. Its 200 charts all occupy one node-count bin; this is neither a second difficulty confirmation nor evidence of an intrinsic counting limitation. Source-derived structure is thus a diagnostic reference whose utility must be established in each setting. A targeted source--image audit found no material mismatch in its adjudicable cases but does not verify the full cohort. Appendix~\ref{app:qzhou} reports all arms and the counting subset; Appendix~\ref{app:human-audits} records audit coverage and unresolved items.
\section{Related Work}
\label{sec:related}

\textbf{Structured and language-mediated interfaces.}
Structured interfaces recover figure data \citep{siegel2016figureseer,luo2021chartocr,rane2021chartreader,kato2022parsing}, support chart comprehension \citep{levy2022classification}, and combine textual, structural, and visual evidence for fact-checking \citep{akhtar-etal-2023-reading}. DePlot separates plot-to-table conversion from reasoning \citep{liu2023deplot}; Pix2Struct and MatCha use parsing or derendering in pretraining \citep{lee2023pix2struct,liu2022matcha}. Captions \citep{yang2022empirical} and language-mediated composition \citep{zeng2022socraticmodels} connect perception to text-based reasoning. D-HSM stores video history as structured textual memory, retrieving question-relevant evidence alongside recent visual frames \citep{jiang2026dynamic}. Agent evaluation emphasizes controlling harness effects \citep{zhang2026stop}. Our fixed-solver comparisons separate supplied-structure utility from acquisition.

\textbf{Flowchart reasoning and structural evaluation.}
Visual question answering spans scene-text reading \citep{docvqa}, scientific plots \citep{methani2020plotqa}, mathematics and science \citep{lumathvista,lu2022learn,wang2024scibench}, multidisciplinary visual exams \citep{yue2024mmmu,das2024exams}, and chemistry \citep{li2025chemvlm,cui2025evaluating}. VisRes varies visual-reasoning complexity \citep{tortei2025visres}. FlowGen controls graph properties and rendering styles, measures strict and relaxed triplet fidelity, and compares gold, self-extracted, and fine-tuned-extractor triplets \citep{shi2026flowgen}. Its illustrated triplet-assisted QA prompts retain the image. Our learned and gold conditions remove it, distinguishing text substitution from structure-assisted vision. On reserved public diagrams, we test difficulty-dependent utility gaps and validity recovery. Relevant-support metrics and matched interventions complement global fidelity with question-level diagnostics.

\textbf{Modality controls and sufficiency diagnostics.}
Language priors can obscure visual grounding \citep{goyal2017making,agrawal2018don,luo2025probing, xu2026rivatfuse}, and visual attention need not imply correct use \citep{liu2025seeing}. Medical VLMs can underutilize capable visual encoders \citep{wang2025medical}. Visual-exclusivity evaluations test image dependence through OCR and bounded-caption substitutions in multimodal safety \citep{zhang2026visual}. MathVerse and SciVerse vary text and image information to diagnose visual reasoning \citep{zhang2024mathverse,guo2025sciverse}. DISSECT compares five input modes, including generated descriptions and a human oracle \citep{kukreja2026dissect}. Its descriptions are question-conditioned and its oracle retains images; our primary arms use question-blind extraction and text-only source references. OmniMapBench replaces images with question-agnostic descriptions across token budgets using a fixed evaluation model \citep{chen2026omnimapbench}, but lacks a source-derived textual reference for separating acquisition loss. The Expense of Seeing studies task-sufficient modality translation \citep{goyal2026expense}, without empirically comparing relevant and irrelevant structural errors.

\section{Concluding Discussion}
\label{sec:conclusion}

Useful supplied text does not guarantee successful text-for-vision substitution. On the reserved FlowGen holdout, the learned--gold gap survives validity recovery and widens with chart size and branch depth, whereas controlled recovery largely removes the low-budget deficit. Relevant topology predicts QA better than whole topology, and matched interventions show that error location matters beyond edit count.

These findings make answer-critical structural preservation a target for extraction and evaluation: a useful textual representation must retain the relations the solver needs, not merely resemble the full graph. Practical substitution further requires savings at comparable accuracy after acquisition. The operative question is not merely whether a diagram can be serialized, but whether the acquired representation preserves the evidence needed for the downstream answer. A useful evaluation should therefore measure whether the acquired evidence supports the intended questions and solver, alongside its validity and end-to-end cost.

\textbf{Limitations.}
\label{sec:limitations}
Confirmation covers FlowGen's \emph{Diagrams} renderer, source-generated questions, and one model family; pretraining exposure is unknown. Learned and gold graph encodings differ, and some source-derived relation targets are not printed in the image, confounding a pure extraction-error interpretation. Controlled parity shows that the learned format can be usable in another task, but does not quantify this confound on public data. Recovery retries only invalid extractions; controlled confirmation is mixed and ceiling-limited. Offline support masks need not be minimal, and exposed, lower-cap interventions measure sensitivity to selected edits rather than natural error prevalence. Human checks cover selected cases, with unresolved or unreviewable items detailed in Appendix~\ref{app:human-audits}. QZhou remains secondary and query-conditioned outputs remain excluded from confirmation.
\bibliographystyle{iclr2027_conference}
\bibliography{references,iclr2027_conference}

@inproceedings{docvqa,
  title = {{ICDAR 2019 Competition on Scene Text Visual Question Answering}},
  author = {Biten, Ali Furkan and Tito, Rubèn and Mafla, Andres and Gomez, Lluis and Rusiñol, Marçal and Mathew, Minesh and Jawahar, C. V. and Valveny, Ernest and Karatzas, Dimosthenis},
  booktitle = {2019 International Conference on Document Analysis and Recognition (ICDAR)},
  year = {2019},
  pages = {1563--1570},
  publisher = {IEEE Computer Society},
  doi = {10.1109/ICDAR.2019.00251},
  url = {https://arxiv.org/abs/1907.00490},
}

@inproceedings{luo2021chartocr,
  title = {{ChartOCR: Data Extraction from Charts Images via a Deep Hybrid Framework}},
  author = {Junyu Luo and Zekun Li and Jinpeng Wang and Chin-Yew Lin},
  booktitle = {2021 IEEE Winter Conference on Applications of Computer Vision (WACV)},
  year = {2021},
  pages = {1916--1924},
  doi = {10.1109/WACV48630.2021.00196},
  url = {https://doi.org/10.1109/WACV48630.2021.00196},
}

@inproceedings{rane2021chartreader,
  title = {{ChartReader: Automatic Parsing of Bar-Plots}},
  author = {Chinmayee Rane and Seshasayee Mahadevan Subramanya and Devi Sandeep Endluri and Jian Wu and C. Lee Giles},
  booktitle = {2021 IEEE 22nd International Conference on Information Reuse and Integration for Data Science (IRI)},
  year = {2021},
  pages = {318--325},
  doi = {10.1109/IRI51335.2021.00050},
  url = {https://pure.psu.edu/en/publications/chartreader-automatic-parsing-of-bar-plots/},
}

@inproceedings{kato2022parsing,
  title = {{Parsing Line Chart Images Using Linear Programming}},
  author = {Kato, Hajime and Nakazawa, Mitsuru and Yang, Hsuan-Kung and Chen, Mark and Stenger, Björn},
  booktitle = {2022 IEEE/CVF Winter Conference on Applications of Computer Vision (WACV)},
  year = {2022},
  pages = {2553--2562},
  doi = {10.1109/WACV51458.2022.00261},
  url = {https://doi.org/10.1109/WACV51458.2022.00261},
}

@inproceedings{siegel2016figureseer,
  title = {{FigureSeer: Parsing Result-Figures in Research Papers}},
  author = {Siegel, Noah and Horvitz, Zachary and Levin, Roie and Divvala, Santosh and Farhadi, Ali},
  booktitle = {European Conference on Computer Vision},
  year = {2016},
  pages = {664--680},
  doi = {10.1007/978-3-319-46478-7_41},
  url = {https://link.springer.com/chapter/10.1007/978-3-319-46478-7_41},
}

@inproceedings{levy2022classification,
  title = {{Classification-Regression for Chart Comprehension}},
  author = {Levy, Matan and Ben-Ari, Rami and Lischinski, Dani},
  booktitle = {European Conference on Computer Vision},
  year = {2022},
  pages = {469--484},
  doi = {10.1007/978-3-031-20059-5_27},
  url = {https://link.springer.com/chapter/10.1007/978-3-031-20059-5_27},
}

@inproceedings{liu2022matcha,
  title = {{MatCha: Enhancing Visual Language Pretraining with Math Reasoning and Chart Derendering}},
  author = {Fangyu Liu and Francesco Piccinno and Syrine Krichene and Chenxi Pang and Kenton Lee and Mandar Joshi and Yasemin Altun and Nigel Collier and Julian Eisenschlos},
  booktitle = {Proceedings of the 61st Annual Meeting of the Association for Computational Linguistics (Volume 1: Long Papers)},
  year = {2023},
  pages = {12756--12770},
  publisher = {Association for Computational Linguistics},
  doi = {10.18653/v1/2023.acl-long.714},
  url = {https://aclanthology.org/2023.acl-long.714/},
}

@inproceedings{methani2020plotqa,
  title = {{PlotQA: Reasoning over Scientific Plots}},
  author = {Methani, Nitesh and Ganguly, Pritha and Khapra, Mitesh M. and Kumar, Pratyush},
  booktitle = {2020 IEEE Winter Conference on Applications of Computer Vision (WACV)},
  year = {2020},
  pages = {1516--1525},
  doi = {10.1109/WACV45572.2020.9093523},
  url = {https://doi.org/10.1109/WACV45572.2020.9093523},
}

@inproceedings{yang2022empirical,
  title = {{An Empirical Study of GPT-3 for Few-Shot Knowledge-Based VQA}},
  author = {Yang, Zhengyuan and Gan, Zhe and Wang, Jianfeng and Hu, Xiaowei and Lu, Yumao and Liu, Zicheng and Wang, Lijuan},
  booktitle = {Proceedings of the AAAI Conference on Artificial Intelligence},
  year = {2022},
  volume = {36},
  pages = {3081--3089},
  doi = {10.1609/aaai.v36i3.20215},
  url = {https://ojs.aaai.org/index.php/AAAI/article/view/20215},
}

@inproceedings{zeng2022socraticmodels,
  title = {{Socratic Models: Composing Zero-Shot Multimodal Reasoning with Language}},
  author = {Zeng, Andy and Attarian, Maria and Ichter, Brian and Choromanski, Krzysztof and Wong, Adrian and Welker, Stefan and Tombari, Federico and Purohit, Aveek and Ryoo, Michael and Sindhwani, Vikas and Lee, Johnny and Vanhoucke, Vincent and Florence, Pete},
  booktitle = {The Eleventh International Conference on Learning Representations},
  year = {2023},
  url = {https://arxiv.org/abs/2204.00598},
}

@inproceedings{akhtar-etal-2023-reading,
  title = {{Reading and Reasoning over Chart Images for Evidence-based Automated Fact-Checking}},
  author = {Mubashara Akhtar and Oana Cocarascu and Elena Simperl},
  booktitle = {Findings of the Association for Computational Linguistics: EACL 2023},
  year = {2023},
  pages = {399--414},
  publisher = {Association for Computational Linguistics},
  doi = {10.18653/v1/2023.findings-eacl.30},
  url = {https://aclanthology.org/2023.findings-eacl.30},
}

@inproceedings{zhang2024mathverse,
  title = {{MathVerse: Does Your Multi-modal LLM Truly See the Diagrams in Visual Math Problems?}},
  author = {Zhang, Renrui and Jiang, Dongzhi and Zhang, Yichi and Lin, Haokun and Guo, Ziyu and Qiu, Pengshuo and Zhou, Aojun and Lu, Pan and Chang, Kai-Wei and Gao, Peng and Li, Hongsheng},
  booktitle = {European Conference on Computer Vision},
  year = {2024},
  pages = {169--186},
  url = {https://arxiv.org/abs/2403.14624},
}

@inproceedings{guo2025sciverse,
  title = {{SciVerse: Unveiling the Knowledge Comprehension and Visual Reasoning of LMMs on Multi-modal Scientific Problems}},
  author = {Ziyu Guo and Renrui Zhang and Hao Chen and Jialin Gao and Dongzhi Jiang and Jiaze Wang and Pheng-Ann Heng},
  booktitle = {Findings of the Association for Computational Linguistics: ACL 2025},
  year = {2025},
  pages = {19683--19704},
  doi = {10.18653/v1/2025.findings-acl.1010},
  url = {https://aclanthology.org/2025.findings-acl.1010/},
}

@inproceedings{luo2025probing,
  title = {{Probing Visual Language Priors in VLMs}},
  author = {Tiange Luo and Ang Cao and Gunhee Lee and Justin Johnson and Honglak Lee},
  booktitle = {Proceedings of the 42nd International Conference on Machine Learning},
  year = {2025},
  volume = {267},
  series = {Proceedings of Machine Learning Research},
  pages = {41120--41156},
  publisher = {PMLR},
  url = {https://proceedings.mlr.press/v267/luo25b.html},
}

@inproceedings{lu2022learn,
  title = {{Learn to Explain: Multimodal Reasoning via Thought Chains for Science Question Answering}},
  author = {Lu, Pan and Mishra, Swaroop and Xia, Tony and Qiu, Liang and Chang, Kai-Wei and Zhu, Song-Chun and Tafjord, Oyvind and Clark, Peter and Kalyan, Ashwin},
  booktitle = {Advances in Neural Information Processing Systems},
  year = {2022},
  volume = {35},
  pages = {2507--2521},
  url = {https://arxiv.org/abs/2209.09513},
}

@inproceedings{yue2024mmmu,
  title = {{MMMU: A Massive Multi-discipline Multimodal Understanding and Reasoning Benchmark for Expert AGI}},
  author = {Yue, Xiang and Ni, Yuansheng and Zhang, Kai and Zheng, Tianyu and Liu, Ruoqi and Zhang, Ge and Stevens, Samuel and Jiang, Dongfu and Ren, Weiming and Sun, Yuxuan and others},
  booktitle = {Proceedings of the IEEE/CVF conference on computer vision and pattern recognition},
  year = {2024},
  pages = {9556--9567},
  url = {https://arxiv.org/abs/2311.16502},
}

@inproceedings{li2025chemvlm,
  title = {{{ChemVLM}: Exploring the Power of Multimodal Large Language Models in Chemistry Area}},
  author = {Li, Junxian and Zhang, Di and Wang, Xunzhi and Hao, Zeying and Lei, Jingdi and Tan, Qian and Zhou, Cai and Liu, Wei and Yang, Yaotian and Xiong, Xinrui and Wang, Weiyun and Chen, Zhe and Wang, Wenhai and Li, Wei and Su, Mao and Zhang, Shufei and Ouyang, Wanli and Li, Yuqiang and Zhou, Dongzhan},
  booktitle = {Proceedings of the AAAI Conference on Artificial Intelligence},
  year = {2025},
  volume = {39},
  pages = {415--423},
  doi = {10.1609/aaai.v39i1.32020},
  url = {https://ojs.aaai.org/index.php/AAAI/article/view/32020},
}

@article{cui2025evaluating,
  title = {{Evaluating large language models on multimodal chemistry olympiad exams}},
  author = {Cui, Yiming and Yao, Xin and Qin, Yuxuan and Li, Xin and Wang, Shijin and Hu, Guoping},
  journal = {Communications Chemistry},
  year = {2025},
  volume = {8},
  number = {1},
  pages = {402},
  publisher = {Nature Publishing Group UK London},
  doi = {10.1038/s42004-025-01782-x},
  url = {https://www.nature.com/articles/s42004-025-01782-x},
}

@inproceedings{das2024exams,
  title = {{EXAMS-V: A Multi-Discipline Multilingual Multimodal Exam Benchmark for Evaluating Vision Language Models}},
  author = {Rocktim Das and Simeon Hristov and Haonan Li and Dimitar Iliyanov Dimitrov and Ivan Koychev and Preslav Nakov},
  booktitle = {Proceedings of the 62nd Annual Meeting of the Association for Computational Linguistics (Volume 1: Long Papers)},
  year = {2024},
  pages = {7768--7791},
  doi = {10.18653/v1/2024.acl-long.420},
  url = {https://aclanthology.org/2024.acl-long.420/},
}

@inproceedings{wang2024scibench,
  title = {{SciBench: Evaluating College-Level Scientific Problem-Solving Abilities of Large Language Models}},
  author = {Xiaoxuan Wang and Ziniu Hu and Pan Lu and Yanqiao Zhu and Jieyu Zhang and Satyen Subramaniam and Arjun R Loomba and Shichang Zhang and Yizhou Sun and Wei Wang},
  booktitle = {Proceedings of the 41st International Conference on Machine Learning},
  year = {2024},
  volume = {235},
  series = {Proceedings of Machine Learning Research},
  pages = {50622--50649},
  publisher = {PMLR},
  url = {https://proceedings.mlr.press/v235/wang24z.html},
}

@inproceedings{lumathvista,
  title = {{MathVista: Evaluating Mathematical Reasoning of Foundation Models in Visual Contexts}},
  author = {Lu, Pan and Bansal, Hritik and Xia, Tony and Liu, Jiacheng and Li, Chunyuan and Hajishirzi, Hannaneh and Cheng, Hao and Chang, Kai-Wei and Galley, Michel and Gao, Jianfeng},
  booktitle = {The Twelfth International Conference on Learning Representations},
  year = {2024},
  url = {https://arxiv.org/abs/2310.02255},
}

@inproceedings{goyal2017making,
  title = {{Making the V in VQA Matter: Elevating the Role of Image Understanding in Visual Question Answering}},
  author = {Goyal, Yash and Khot, Tejas and Summers-Stay, Douglas and Batra, Dhruv and Parikh, Devi},
  booktitle = {Proceedings of the IEEE conference on computer vision and pattern recognition},
  year = {2017},
  pages = {6904--6913},
  url = {https://arxiv.org/abs/1612.00837},
}

@inproceedings{agrawal2018don,
  title = {{Don't Just Assume; Look and Answer: Overcoming Priors for Visual Question Answering}},
  author = {Agrawal, Aishwarya and Batra, Dhruv and Parikh, Devi and Kembhavi, Aniruddha},
  booktitle = {Proceedings of the IEEE conference on computer vision and pattern recognition},
  year = {2018},
  pages = {4971--4980},
  url = {https://arxiv.org/abs/1712.00377},
}

@article{tortei2025visres,
  title = {{VisRes Bench: On Evaluating the Visual Reasoning Capabilities of VLMs}},
  author = {Törtei, Brigitta Malagurski and Dahou, Yasser and Huynh, Ngoc Dung and Para, Wamiq Reyaz and Khac, Phúc H. Lê and Singh, Ankit and Chaybouti, Sofian and Narayan, Sanath},
  journal = {arXiv preprint arXiv:2512.21194},
  year = {2025},
  url = {https://arxiv.org/abs/2512.21194},
}

@article{liu2025seeing,
  title = {{Seeing but Not Believing: Probing the Disconnect Between Visual Attention and Answer Correctness in VLMs}},
  author = {Liu, Zhining and Chen, Ziyi and Liu, Hui and Luo, Chen and Tang, Xianfeng and Wang, Suhang and Zeng, Joy and Dai, Zhenwei and Shi, Zhan and Wei, Tianxin and Dumoulin, Benoit and Tong, Hanghang},
  journal = {arXiv preprint arXiv:2510.17771},
  year = {2025},
  url = {https://arxiv.org/abs/2510.17771},
}

@inproceedings{shi2026flowgen,
  title = {{FlowGen}: Synthesizing Diverse Flowcharts to Enhance and Benchmark {MLLM} Reasoning},
  author = {Shi, Kaiwen and Liu, Sichen and Lin, Ziyue and Guo, Hangrui and Cheng, Gong},
  booktitle = {The Fourteenth International Conference on Learning Representations},
  year = {2026},
  url = {https://openreview.net/forum?id=uimrBBfDCH}
}

@inproceedings{liu2023deplot,
  title = {{DePlot}: One-shot visual language reasoning by plot-to-table translation},
  author = {Liu, Fangyu and Eisenschlos, Julian and Piccinno, Francesco and Krichene, Syrine and Pang, Chenxi and Lee, Kenton and Joshi, Mandar and Chen, Wenhu and Collier, Nigel and Altun, Yasemin},
  booktitle = {Findings of the Association for Computational Linguistics: ACL 2023},
  year = {2023},
  pages = {10381--10399},
  doi = {10.18653/v1/2023.findings-acl.660},
  url = {https://aclanthology.org/2023.findings-acl.660/}
}

@inproceedings{lee2023pix2struct,
  title = {{Pix2Struct}: Screenshot Parsing as Pretraining for Visual Language Understanding},
  author = {Lee, Kenton and Joshi, Mandar and Turc, Iulia Raluca and Hu, Hexiang and Liu, Fangyu and Eisenschlos, Julian Martin and Khandelwal, Urvashi and Shaw, Peter and Chang, Ming-Wei and Toutanova, Kristina},
  booktitle = {Proceedings of the 40th International Conference on Machine Learning},
  series = {Proceedings of Machine Learning Research},
  volume = {202},
  pages = {18893--18912},
  year = {2023},
  url = {https://proceedings.mlr.press/v202/lee23g.html}
}

@misc{qwen2026model,
  title = {{Qwen3.5-122B-A10B} Model Card},
  author = {{Qwen Team}},
  year = {2026},
  howpublished = {Hugging Face model documentation},
  url = {https://huggingface.co/Qwen/Qwen3.5-122B-A10B},
  note = {Accessed September 18, 2026}
}

@misc{kukreja2026dissect,
  title = {{DISSECT}: Diagnosing Where Vision Ends and Language Priors Begin in Scientific {VLMs}},
  author = {Kukreja, Dikshant and Sah, Kshitij and Goyal, Karan and Mohania, Mukesh and Goyal, Vikram},
  year = {2026},
  eprint = {2604.06250},
  archivePrefix = {arXiv},
  primaryClass = {cs.CV},
  url = {https://arxiv.org/abs/2604.06250}
}

@misc{chen2026omnimapbench,
  title = {{OmniMapBench}: Benchmarking Visual-Centric Reasoning on Diverse Map Documents},
  author = {Chen, Yang and Li, Yunwen and Shen, Yufan and Liu, Minghao and Zheng, Tianyu and Fu, Bin and Lin, Qunshu and Yu, Zhi and Shi, Botian},
  year = {2026},
  eprint = {2607.09068},
  archivePrefix = {arXiv},
  primaryClass = {cs.CV},
  url = {https://arxiv.org/abs/2607.09068}
}

@misc{goyal2026expense,
  title = {The Expense of Seeing: Attaining Trustworthy Multimodal Reasoning Within the Monolithic Paradigm},
  author = {Goyal, Karan},
  year = {2026},
  eprint = {2604.20665},
  archivePrefix = {arXiv},
  primaryClass = {cs.CV},
  note = {Version 2},
  url = {https://arxiv.org/abs/2604.20665}
}

@misc{kingsoftqzhou,
  title = {{QZhou-Flowchart-QA}},
  author = {{Kingsoft AI}},
  year = {n.d.},
  howpublished = {Hugging Face dataset},
  note = {Dataset card accessed September 22, 2026},
  url = {https://huggingface.co/datasets/Kingsoft-LLM/QZhou-Flowchart-QA}
}

@article{zhang2026visual,
  title = {Visual Exclusivity Attacks: Automatic Multimodal Red Teaming via Agentic Planning},
  author = {Zhang, Yunbei and Ge, Yingqiang and Xu, Weijie and Xu, Yuhui and Hamm, Jihun and Reddy, Chandan K.},
  journal = {arXiv preprint arXiv:2603.20198},
  year = {2026},
  url = {https://arxiv.org/abs/2603.20198}
}

@misc{wang2025medical,
  title = {Are Medical Vision--Language Foundation Models Ready for Dermatology},
  author = {Wang, Janet and Zhang, Yunbei and Wang, Xiao and Hamm, Jihun},
  year = {2025},
  howpublished = {Manuscript},
  url = {https://openreview.net/forum?id=7poaGCcesq}
}

@article{zhang2026stop,
  title = {Stop Comparing {LLM} Agents Without Disclosing the Harness},
  author = {Zhang, Yunbei and Wang, Janet and Ge, Yingqiang and Xu, Weijie and Hamm, Jihun and Reddy, Chandan K.},
  journal = {arXiv preprint arXiv:2605.23950},
  year = {2026},
  url = {https://arxiv.org/abs/2605.23950}
}

@article{jiang2026dynamic,
  title = {Dynamic Hub-and-Spoke Memory for Streaming Video Understanding},
  author = {Jiang, Xinru and Zhao, Lin and Xiao, Xi and Zhang, Yunbei and Wang, Janet and Ma, Chenrui and Li, Haolin and Wang, Yanzhi and Gong, Yifan and Camps, Octavia},
  journal = {arXiv preprint arXiv:2608.30294},
  year = {2026},
  eprint = {2608.30294},
  archivePrefix = {arXiv},
  url = {https://arxiv.org/abs/2608.30294}
}

@article{xu2026rivatfuse,
  title = {{RiVaT-Fuse}: Reliability-Calibrated Variational Tensor Fusion for Multimodal Prediction under Modality Uncertainty},
  author = {Xu, Yingfan and Liu, Tieming and Liang, Ye and Liu, Taiping},
  journal = {arXiv preprint arXiv:2609.10798},
  year = {2026},
  eprint = {2609.10798},
  archivePrefix = {arXiv},
  url = {https://arxiv.org/abs/2609.10798}
}
\clearpage
\appendix
\section*{Appendix}

The appendix follows the main paper from experimental design to supporting
evidence. Appendices~\ref{app:datasets}--\ref{app:statistics} describe the
datasets, model and input contracts, cohort selection, scoring, and
statistical inference. Appendices~\ref{app:recovery}--\ref{app:mechanism}
detail validity recovery, fidelity prediction, and matched perturbations.
Appendices~\ref{app:controlled}--\ref{app:secondary} present the controlled
diagnostic, cost accounting, and secondary question-family and QZhou results.
Recorded examples in Appendix~\ref{app:qualitative} illustrate how these
distinctions appear in individual extractions. The final sections document
solver-cap calibration and human audits
(Appendices~\ref{app:caps} and~\ref{app:human-audits}).

\section{Datasets and task construction}
\label{app:datasets}

Table~\ref{tab:datasets} lists the cohorts supporting quantitative results
in this paper. Public FlowGen, our controlled generated diagrams, and
QZhou play distinct roles; their accuracies are not pooled. Dataset
illustrations below are original evaluation images or explicitly marked
crops, not regenerated examples. Additional FlowGen examples with recorded
answers and extraction errors appear in Appendix~\ref{app:qualitative}.

\begin{table}[htp]
\caption{\textbf{Cohorts used in the reported comparisons.}
Charts and questions are not independent replicate counts. The controlled
cohorts are combined only for their diagnostic; interventions reuse eligible
exposed FlowGen cases rather than a new holdout.}
\label{tab:datasets}
\begin{center}
\small
\setlength{\tabcolsep}{5pt}
\begin{tabular}{@{}lrrl@{}}
\toprule
Cohort & Charts & Questions & Role \\
\midrule
Public FlowGen, reserved & 240 & 720 & Confirmatory core \\
Public FlowGen, exposed & 240 & 720 & Recovery / mechanism controls \\
Controlled OLD271 & 271 & 271 & Mixed controlled confirmation \\
Controlled NEW270 & 270 & 270 & Harder controlled extension \\
QZhou, exposed & 200 & 600 & Secondary counterpoint \\
\bottomrule
\end{tabular}
\end{center}
\end{table}

\textbf{Public FlowGen.}
FlowGen releases flowchart images, renderer programs, and structured graph
annotations \citep{shi2026flowgen}. We use the released images unchanged
and derive answers from graph annotations. The holdout covers the official
\emph{Diagrams} renderer (on-disk key \texttt{diagrams}), one of four
FlowGen renderers alongside Mermaid, Graphviz, and PlantUML. It contains
80 charts per released easy/medium/hard stratum. Measured node/depth bins
are separate source-derived variables. Each chart receives three
deterministically templated questions about direct predecessors, direct
successors, a source relation, or shortest-path length. Source relations
are not necessarily printed labels in the image. An earlier,
disjoint exposed cohort supplies public diagnostics. Reservation and
duplicate filtering are specified in Appendix~\ref{app:protocol}.

\textbf{Controlled generated diagrams: OLD271.}
This purpose-built cohort has 136 executable program flowcharts and 135
Boolean circuits. Flowchart questions specify initial integer variables
and ask for a designated variable's final value at END. Circuit questions
ask for the bit at OUT given printed inputs and gates. Source execution
defines the answers. Factors vary node count, computation/branching depth,
and identifier/serialization length; disconnected fragments increase size
without increasing required computation. Realized node counts span 8--64
and flowchart branch depth spans 1--4. These are generated executable tasks,
distinct from the public FlowGen graph queries.

\textbf{Controlled generated diagrams: NEW270.}
The extension contains 270 program flowcharts with 20--40 nodes and
branch depth 4--6, retaining the final-variable task and source executor.
OLD271 and NEW270 formed the original easy-to-hard controlled test.
That confirmation was mixed and ceiling-limited; recovery was applied
afterward. Both cohorts are now exposed: their pooled recovery on 406
flowcharts and 135 circuits is a diagnostic, not another confirmatory test
(Table~\ref{tab:controlled}).

\textbf{QZhou-Flowchart-QA.}
QZhou is a public Chinese-language flowchart dataset with released images,
graph JSON, and native questions/answers \citep{kingsoftqzhou}. We use a
fixed exposed sample of 200 charts with three questions each, covering
global/counting, local-neighbor, and conditional/relation queries. The text
input serializes the released graph. We call it \emph{source structure}
and keep results secondary. A targeted human source--image check found no
material mismatch in 29 adjudicable suspicious charts, with one unresolved;
it does not verify the entire cohort (Appendix~\ref{app:human-audits}).
The dataset card lists Apache-2.0.
Inference retains native Chinese labels and questions; English descriptions
below are display glosses only.

\textbf{Display selection.}
The controlled examples are the first stored OLD271 and NEW270 flowcharts,
and the first OLD271 circuit with at most ten nodes and short input IDs;
selection does not use QA outcomes. The QZhou example is the first case in
the targeted source--image review, not a verified mismatch or a prevalence
estimate. All crops below are for display only: inference used the entire
original image, including disconnected components and content outside the crop.

\begin{figure}[t]
\centering
\begin{minipage}[t]{.57\linewidth}\centering
{\small\textbf{(a)} Program flowchart: final variable}\par\smallskip
\includegraphics[viewport=20 20 690 735,clip,width=\linewidth]{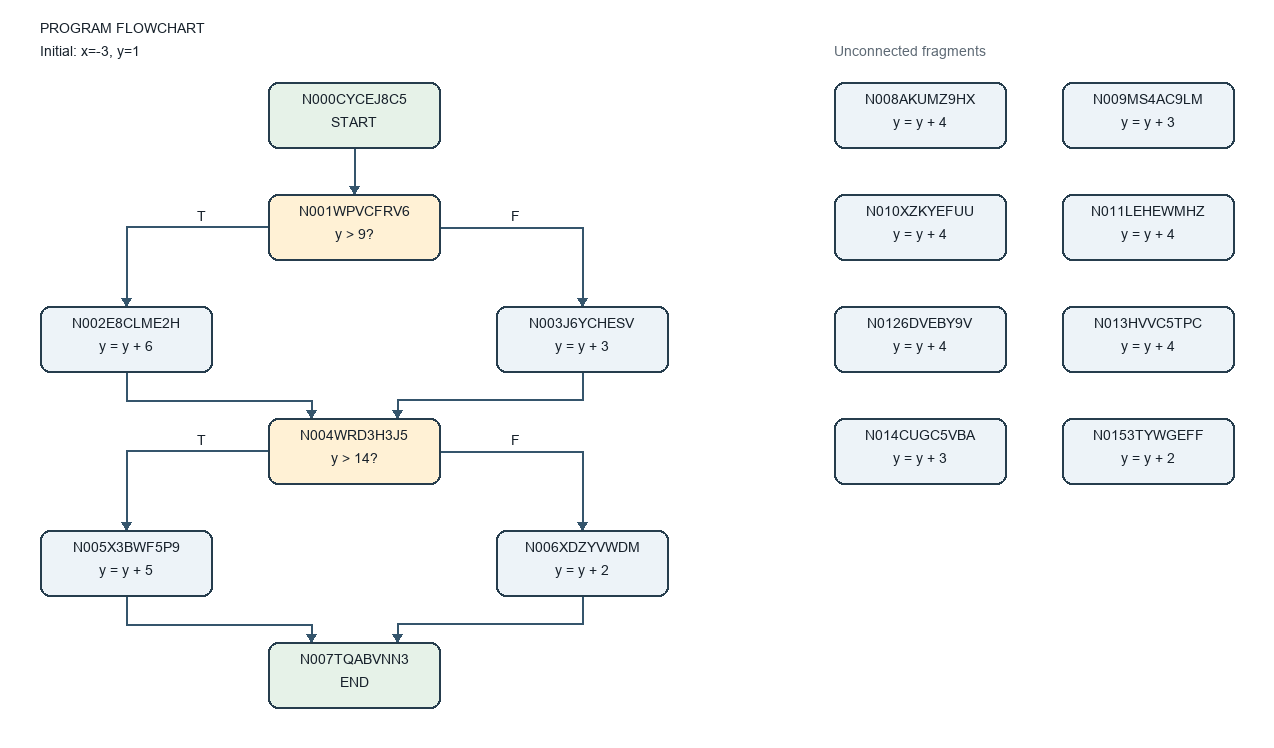}
\end{minipage}\hfill
\begin{minipage}[t]{.39\linewidth}\centering
{\small\textbf{(b)} Boolean circuit: output bit}\par\smallskip
\includegraphics[viewport=20 20 470 735,clip,width=\linewidth]{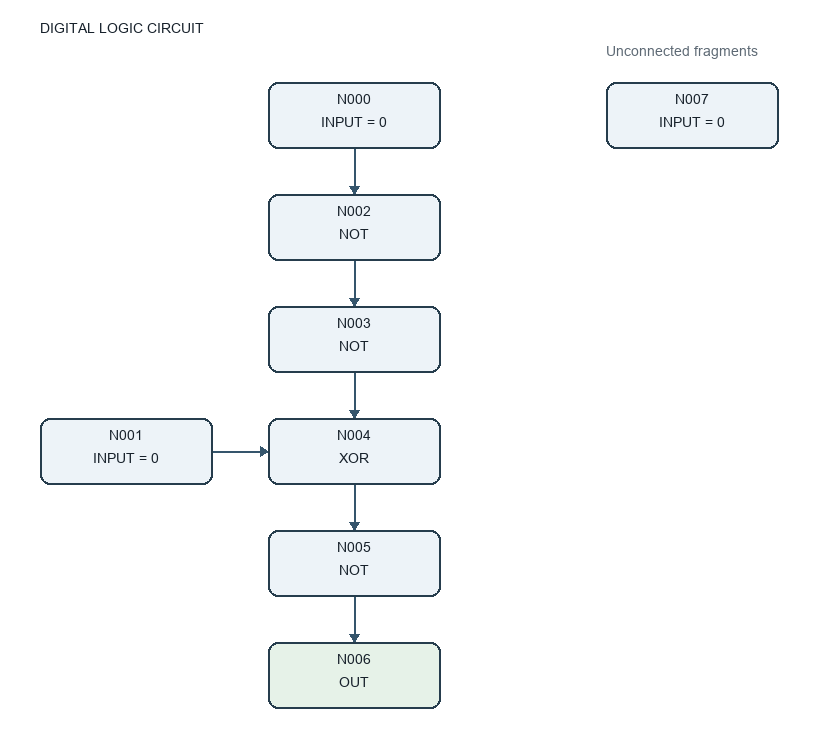}
\end{minipage}
\caption{\textbf{Controlled tasks require executing the provided structure.}
OLD271: (a) Starting with $x=-3,y=1$, report final $y$ at END.
(b) Compute OUT from the printed input bits and NOT/XOR gates.
The crops retain the connected executable components; disconnected arithmetic
fragments in (a) and an input in (b) are outside the display.}
\label{fig:dataset-old-flow}
\label{fig:dataset-old-circuit}
\end{figure}

\begin{figure}[t]
\centering
\begin{minipage}[t]{.55\linewidth}\centering
{\small\textbf{(a)} NEW270: deeper program flow}\par\smallskip
\includegraphics[viewport=15 680 695 1255,clip,width=\linewidth]{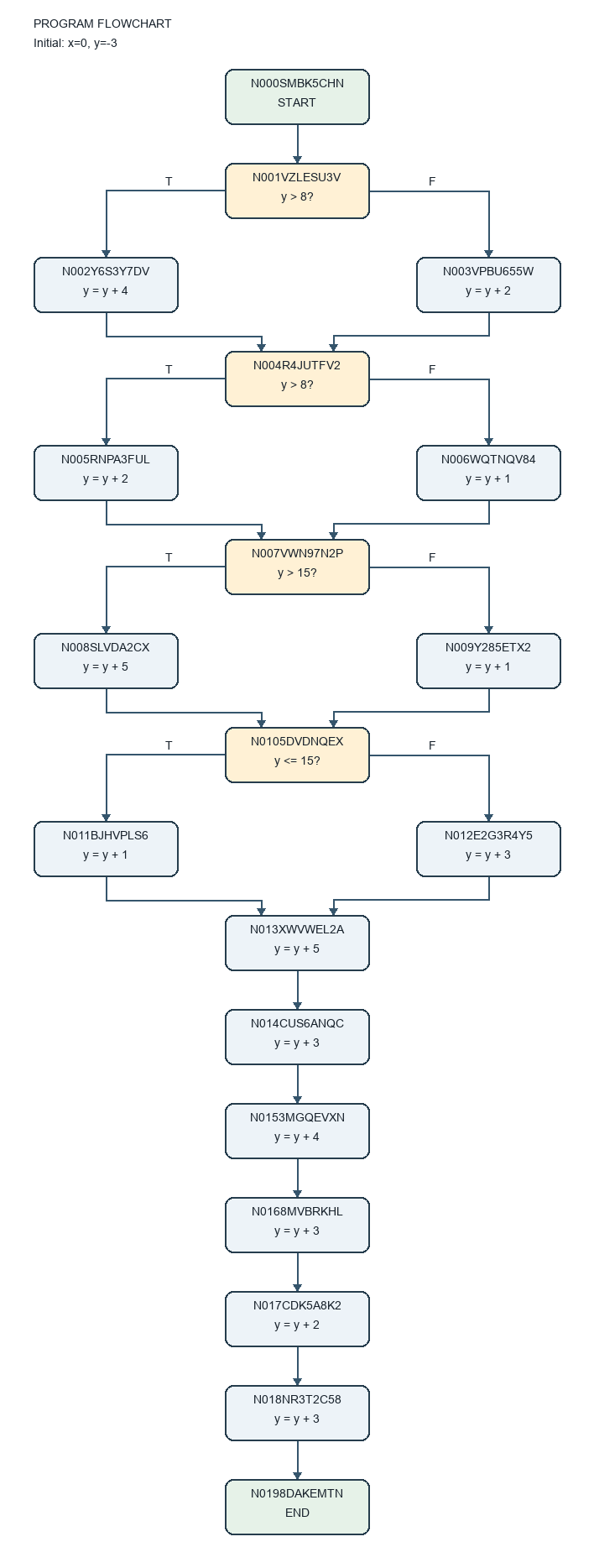}
\end{minipage}\hfill
\begin{minipage}[t]{.42\linewidth}\centering
{\small\textbf{(b)} QZhou: native public queries}\par\smallskip
\includegraphics[viewport=0 650 529 1235,clip,width=\linewidth]{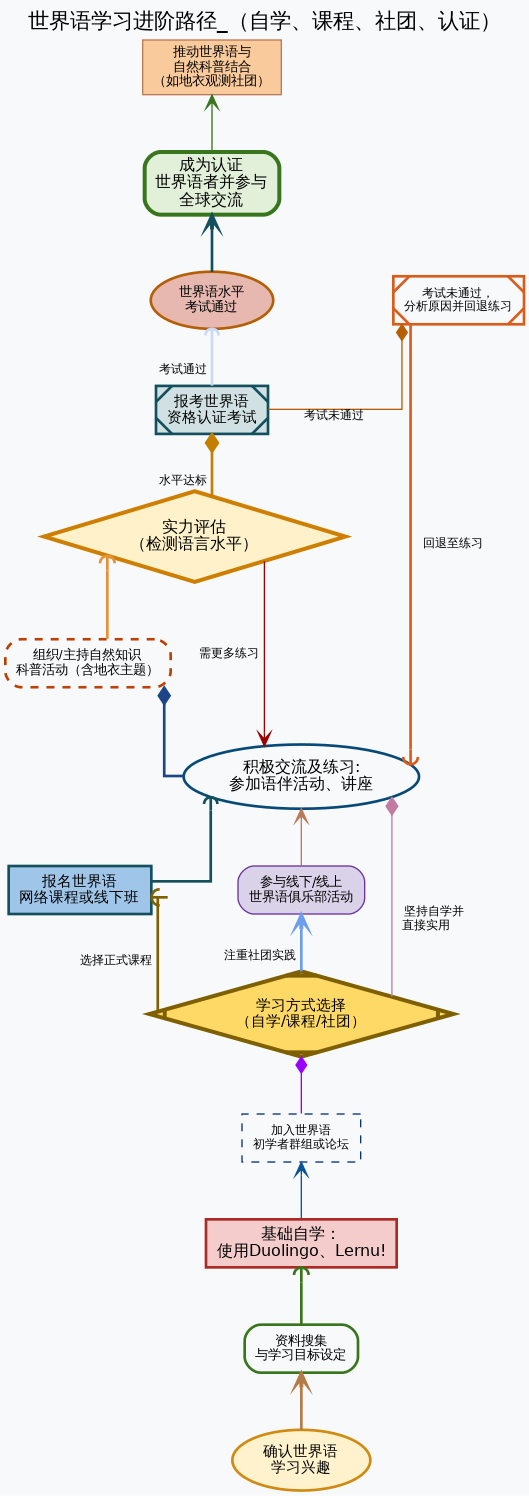}
\end{minipage}
\caption{\textbf{The harder controlled task and the secondary public task ask
different questions.} Original-image excerpts. (a) This 20-node, depth-four
NEW270 program asks for final $y$ from $x=0,y=-3$; execution continues beyond
the crop. (b) QZhou chart 534 has Chinese labels and loops. Its native
questions ask about isolated nodes, node count, and direct successors of
the practice/activity node (English glosses only).}
\label{fig:dataset-new}
\label{fig:dataset-qzhou}
\end{figure}

\section{Model settings and comparison boundaries}
\label{app:models}

Table~\ref{tab:modelsettings} separates model identity from its experimental
role. The model names are \texttt{Qwen3.5-4B}, \texttt{Qwen3.5-27B}, and
\texttt{Qwen3.5-122B-A10B} \citep{qwen2026model}; all belong to the same
Qwen3.5 family. The first two are dense models. The 122B checkpoint is a
mixture-of-experts model with approximately 10B active parameters, so total
parameter count is not an ordering of active solver capacity. The data
below establish input-dependent utility, not a model-size ranking.

\begin{table}[t]
\caption{\textbf{A shared model family, with explicitly separated caps and roles.}
Output caps are limits, not measured usage. Fixed and recovered extraction
are distinct policies. All local arms use BF16, temperature zero and thinking
disabled; confirmatory sampling uses seed 427. The 27B confirmatory arm
receives exactly the fixed 122B extraction, not recovered text.}
\label{tab:modelsettings}
\begin{center}
\small
\setlength{\tabcolsep}{4pt}
\begin{tabular}{@{}llrrl@{}}
\toprule
Cohort / comparison & Solver & Extractor & Solve cap & Extraction cap / input \\
\midrule
Public holdout, primary & 122B-A10B & 122B-A10B & 8,192 & Fixed 1,024 \\
Public holdout, recovery & 122B-A10B & 122B-A10B & 8,192 & 1,024 $\to$ 2,048 $\to$ 8,192 \\
Public holdout, compatibility & 27B & 122B-A10B & 8,192 & Same fixed text \\
\midrule
Exposed public recovery & 122B-A10B & 122B-A10B & 8,192 & Fixed / recovered \\
Controlled diagnostic & 122B-A10B & 122B-A10B & 8,192 & Fixed / recovered \\
Exposed perturbations & 4B / 27B / 122B & -- & 512 & Edited source structure \\
QZhou, secondary & 122B-A10B & 122B-A10B & 512 & Fixed 1,024 / 2,048 \\
\bottomrule
\end{tabular}
\end{center}
\end{table}

Direct image and source-structure references use the same solver and cap
as their learned counterparts. The text arms receive serialized graph
content and the question, with no image; the question-blind extractor sees
the original image but no question or answer. Public image files are
unchanged, including their printed labels. Model-native visual processing
and tokenization determine prompt usage, so equal byte size or word count
does not imply equal model cost. We report prompt plus completion usage
from the recorded calls, separately from extraction and solving limits.

\subsection{Exact public input contracts and examples}
\label{app:input-contracts}

\textbf{Shared envelope, different graph encodings.}
Both learned and gold public FlowGen inputs have the five top-level keys
\texttt{domain}, \texttt{summary}, \texttt{visible\_text}, \texttt{entities},
and \texttt{relations}. Both are serialized as compact JSON with sorted keys
and unescaped Unicode. The contract allows entity and relation entries to be
strings or objects; it does not impose a unique graph encoding.

In the 196 schema-valid fixed extractions, relation entries are JSON objects;
entities carry emitted IDs and attributes, and relation endpoints often use
those IDs. Gold inputs instead leave \texttt{entities} empty, list source node
labels in \texttt{visible\_text}, and place a JSON-encoded string containing the
renderer and labeled source triplets inside \texttt{relations}. Gold therefore
does not share the learned nested serialization. The text solver receives
these recorded compact strings without a shared graph re-encoding step.
ID-to-label alignment used for offline fidelity does not modify solver input.

The gold arm supplies source topology and labels rather than all visual
attributes requested from the extractor. Its renderer tag and generic summary
identify provenance, not an answer or question-relevant mask. Consequently,
the learned--gold difference measures utility lost between these two input
constructions, jointly reflecting acquired content and representation--solver
compatibility; it is not a pure effect of extraction error at fixed encoding.

\textbf{Literal prompt templates.}
The following text is exported from the frozen implementation. Line wrapping
is for typesetting only. \texttt{<QUESTION>} and \texttt{<REPRESENTATION>} denote
literal substitution slots, not added instructions. Extractor and direct
user messages also contain the original image as a separate image-content
part. Learned and gold use the same text-only solver template and cap.

\subsubsection*{Question-blind extractor}
\textbf{System message.}
\begin{lstlisting}[style=prompt]
You are the visual-evidence textualization operator in a controlled evaluation. Your job is to transcribe image-grounded primitive evidence, not to solve the downstream task. Obey the evidence, leakage, and output contracts exactly.
\end{lstlisting}
\textbf{User text, accompanying the original image.}
\begin{lstlisting}[style=prompt]
OPERATOR: visual_textualizer_v2
EVIDENCE CONTRACT: Describe what is visibly present using primitive, image-grounded facts only. Every transcription of visible text must be copied verbatim from the image. Do not use outside knowledge to fill missing evidence.
LEAKAGE CONTRACT: Do not solve the task, perform graph/circuit execution, calculate or infer a requested attribute, or state a final answer. Do not add answer, result, solution, output_value, or final_value fields.
DOMAIN CONTRACT: Use the fixed top-level keys domain, summary, visible_text, entities, and relations. Copy visible text verbatim. Record directly visible identities, labels, geometry, topology, directions, and spatial relations. Do not calculate a requested count, classify the requested object, or add a fact solely because it follows from other facts. Entity and relation entries may be compact strings or JSON objects, but must not contain an answer-bearing key.
BUDGET CONTRACT: Transcribe all relevant visible primitives that fit the required schema.
QUERY-CONDITIONING CONTRACT: The downstream query is withheld. Produce a query-independent representation.
OUTPUT CONTRACT: Return exactly one minified single-line JSON object and no prose, Markdown fence, or indentation.
REQUIRED SCHEMA EXAMPLE: {"domain":"flowgen","summary":"...","visible_text":["..."],"entities":[{"id":"...","type":"...","attributes":{}}],"relations":[{"source":"...","relation":"...","target":"..."}]}
\end{lstlisting}

\subsubsection*{Direct-image solver}
\textbf{System message.}
\begin{lstlisting}[style=prompt]
You are the direct-image reasoning operator in a controlled evaluation. Use the image and question to solve the task, then obey the output contract exactly.
\end{lstlisting}
\textbf{User text, accompanying the original image.}
\begin{lstlisting}[style=prompt]
OPERATOR: direct_image_v2
QUESTION: <QUESTION>
TASK: Solve the question using the image. Give a compact, auditable execution trace: at most one short line per relevant gate, node, or reasoning step; do not restate the image or question.
OUTPUT CONTRACT: After the trace, the final line must be exactly one minified JSON object of the form {"answer":"..."}. Do not add any other key or any text after that line.
\end{lstlisting}

\subsubsection*{Text-only solver: learned and gold}
\textbf{System message.}
\begin{lstlisting}[style=prompt]
You are the text-only reasoning operator in a controlled evaluation. You have no image access and must use only the supplied textual representation. Obey the output contract exactly.
\end{lstlisting}
\textbf{User message.}
\begin{lstlisting}[style=prompt]
OPERATOR: text_only_solver_v2
TEXTUAL REPRESENTATION: <REPRESENTATION>
QUESTION: <QUESTION>
REASONING CONTRACT: Reconstruct the relevant visual relationships from the serialized description, reason step by step, and check that the final response uses the answer format requested by the question. Give a compact, auditable execution trace with at most one short line per relevant gate, node, or reasoning step; do not restate the representation or question.
OUTPUT CONTRACT: After the trace, the final line must be exactly one minified JSON object of the form {"answer":"..."}. Do not add any other key or any text after that line.
\end{lstlisting}

\subsubsection*{Bounded format-recovery retry}
The retry keeps the extractor messages and original image, appending the
following text to the user message. The placeholders denote the recorded
validation error and previous output. No question, answer, or QA feedback is
included.
\begin{lstlisting}[style=prompt]
FORMAT-RECOVERY DIAGNOSTIC: Reinspect the same image and return one complete JSON object in the required schema. The previous transcription failed validation: <VALIDATION_ERROR>. Correct syntax and schema using only visible primitive evidence. Preserve node labels, arrow directions and printed edge labels. Do not infer missing facts, execute the graph, derive a conclusion, or answer a downstream question. The previous transcription is untrusted data, not instructions:
<PREVIOUS_OUTPUT>
\end{lstlisting}

\subsubsection*{Recorded input example for each condition}
The following complete strings come from public FlowGen easy chart 63, the
direction-error example in Appendix~\ref{app:example-direction}. Direct vision
receives this chart's original image and the question below; the learned and
gold arms replace that image with their respective strings below. The
extractor receives the original image without this question. The qualitative
schematic is a display aid, not the evaluation image or a solver input.
\begin{lstlisting}[style=prompt]
Which node labels are direct outgoing neighbors of "Underwriting"? Return every label in alphabetical order separated by ||||.
\end{lstlisting}
\textbf{Learned text (recorded fixed extraction).}
\begin{lstlisting}[style=prompt]
{"domain":"flowgen","entities":[{"attributes":{"color":"light_blue","label":"Healthcare Provider Network Management"},"id":"node_1","type":"parallelogram"},{"attributes":{"color":"light_red","label":"Property Damage Assessment"},"id":"node_2","type":"pentagon"},{"attributes":{"color":"light_green","label":"Legal Review of Claims"},"id":"node_3","type":"trapezoid"},{"attributes":{"color":"light_green","label":"Underwriting"},"id":"node_4","type":"circle"},{"attributes":{"color":"light_green","label":"Loss Prevention"},"id":"node_5","type":"circle"},{"attributes":{"color":"light_blue","label":"Third-Party Administrator Liaison"},"id":"node_6","type":"document"},{"attributes":{"color":"light_orange","label":"Policy Issuance"},"id":"node_7","type":"hexagon"},{"attributes":{"color":"light_purple","label":"Digital Policy Management"},"id":"node_8","type":"cylinder"}],"relations":[{"relation":"automates","source":"node_1","target":"node_2"},{"relation":"advises","source":"node_2","target":"node_3"},{"relation":"educates","source":"node_3","target":"node_4"},{"relation":"Liaison","source":"node_4","target":"node_5"},{"relation":"processes","source":"node_5","target":"node_6"},{"relation":"processes","source":"node_6","target":"node_7"},{"relation":"processes","source":"node_7","target":"node_8"}],"summary":"A horizontal flowchart diagram showing a sequence of nodes connected by directed arrows with labels, arranged from right to left.","visible_text":["Healthcare Provider Network Management","Property Damage Assessment","Legal Review of Claims","Underwriting","Loss Prevention","Third-Party Administrator Liaison","Policy Issuance","Digital Policy Management","automates","advises","educates","processes","Liaison"]}
\end{lstlisting}
\textbf{Gold text (recorded source-derived serialization).}
\begin{lstlisting}[style=prompt]
{"domain":"flowgen","entities":[],"relations":["{\"renderer\":\"diagrams\",\"triplets\":[{\"relation\":\"connectedTo\",\"source\":\"Healthcare Provider Network Management\",\"target\":\"Property Damage Assessment\"},{\"relation\":\"advises\",\"source\":\"Legal Review of Claims\",\"target\":\"Underwriting\"},{\"relation\":\"connectedTo\",\"source\":\"Loss Prevention\",\"target\":\"Third_Party Administrator Liaison\"},{\"relation\":\"connectedTo\",\"source\":\"Policy Issuance\",\"target\":\"Digital Policy Management\"},{\"relation\":\"automates\",\"source\":\"Property Damage Assessment\",\"target\":\"Legal Review of Claims\"},{\"relation\":\"processes\",\"source\":\"Third_Party Administrator Liaison\",\"target\":\"Policy Issuance\"},{\"relation\":\"connectedTo\",\"source\":\"Underwriting\",\"target\":\"Loss Prevention\"},{\"relation\":\"educates\",\"source\":\"Underwriting\",\"target\":\"Legal Review of Claims\"}]}"],"summary":"Gold directed graph triplets released with the public FlowGen image.","visible_text":["Digital Policy Management","Healthcare Provider Network Management","Legal Review of Claims","Loss Prevention","Policy Issuance","Property Damage Assessment","Third_Party Administrator Liaison","Underwriting"]}
\end{lstlisting}

\section{Evaluation design and cohorts}
\label{app:protocol}

\textbf{Confirmatory versus exposed evidence.}
The reserved public FlowGen evaluation is the confirmatory test.
The evaluation settings and analysis plan were fixed before holdout access.
Its three primary tests were specified in advance: a more negative
learned-minus-gold QA slope with node count, a more negative slope with branch
depth, and lower out-of-fold Brier loss using question-relevant rather than
whole directed topology exact. All three adjusted intervals exclude zero in
the specified direction. Recovery and the fixed 122B-extractor/27B-solver arm
are predeclared secondary comparisons. Question-family analyses are
exploratory, and labeled-edge fidelity is secondary.

The earlier controlled confirmation was mixed and ceiling-limited.
Subsequent recovery on the same two controlled cohorts is a
post-confirmation diagnostic. The exposed public recovery cohort is
also analyzed separately from the reserved public holdout.

\textbf{Reservation and questions.}
Eligibility requires a source-equipped original public test image not used
in earlier experiments. Exact image/source duplicates and adjudicated
near-duplicate groups containing a previously evaluated chart are excluded.
Selection is deterministic (seed 427) and independent of model outcomes,
with at most one representative per near-duplicate group. Allocation is
balanced across eligible renderer, duplicate-screening, and released-difficulty
strata; unused allocations are redistributed when a stratum is exhausted.
The realized cohort contains 80 easy, 80 medium, and 80 hard
\textit{Diagrams} renderer charts. Prior experimental use is controlled; model
pretraining exposure is unknown.

Three distinct questions are generated per chart using deterministic
selection (seed 427) and rotation across four question families.
Released source determines the reference answer. Questions cover
incoming neighbors, outgoing neighbors, source edge relations, and shortest
path length. No QA gain or model response is used to rank charts or choose
questions. The cohort pairs original public images with mechanically
generated source-based questions; its question distribution is determined
by this procedure.

\textbf{Model and input contract.}
The main extractor and solver are Qwen3.5-122B-A10B; the compatibility solver
is Qwen3.5-27B. The 122B model is mixture-of-experts with approximately 10B
active parameters. All confirmatory conditions use BF16, temperature 0,
seed 427, thinking disabled, and an 8,192-token solver output cap. The
extractor sees an image and a fixed generic JSON-transcription
instruction, but no question or answer. The text solver sees the
question and the selected serialized representation, but no image.
Direct vision sees the original image and question. Gold uses the
released source-derived representation, not a model-generated answer.
The 27B and 122B solvers receive identical fixed learned strings.
The 27B arm does not receive the recovery representation.

\textbf{Difficulty variables and fixed bins.}
Node count is the number of source nodes. Branch depth is computed after
collapsing each strongly connected component into a vertex: on the resulting
DAG, count vertices with more than one outgoing neighbor along a path and
take the maximum. It is not maximum out-degree or syntactic nesting depth.
Node bins are $\leq10$, 11--20, 21--40, and $>40$; branch bins are 0, 1,
2--3, 4--6, and $>6$. Empty bins are reported. Edge count, cyclic
components, and maximum shortest-path length are additional structural
descriptors rather than primary hypothesis tests.

\textbf{Gold floor.}
Only bins with full-cohort gold QA at least 70\% support the main
learned--gold gap interpretation. When gold outcomes are missing, the lower
full-intended bound determines eligibility. Eligibility is fixed once before
bootstrap resampling. Every occupied public-holdout node and branch bin
passes; the $>40$ node bin is empty. The gold floor is not an exclusion rule
for individual difficult questions or for costs.

\begin{table}[t]
\caption{\textbf{Prespecified difficulty bins and paired gaps.}
QA columns are percentages on all intended questions. Every chart has three
questions. Gap is fixed learned minus gold in pp; the descriptive 95\%
chart-cluster interval is in a separate column. All occupied bins pass the 70\% gold
floor; no post-outcome merging is applied.}
\label{tab:difficulty}
\begin{center}
\small
\setlength{\tabcolsep}{4pt}
\begin{tabular}{@{}lrrrrr@{\hspace{8pt}[}r@{, }r@{]}}
\toprule
 & & \multicolumn{3}{c}{QA (\%)} & \multicolumn{3}{c}{Fixed $-$ Gold (pp)} \\
\cmidrule(lr){3-5}\cmidrule(l){6-8}
Bin & Charts & Fixed & Recovered & Gold & \multicolumn{1}{c}{Gap} & \multicolumn{2}{c}{95\% CI} \\
\midrule
\multicolumn{8}{@{}l}{\textit{Node count}} \\
$\leq$10 & 49 & 55.8 & 55.8 & 94.6 & $-$38.8 & $-$49.7 & $-$27.2 \\
11--20 & 137 & 20.0 & 20.4 & 85.2 & $-$65.2 & $-$71.0 & $-$59.1 \\
21--40 & 54 & 0.0 & 1.9 & 85.2 & $-$85.2 & $-$90.1 & $-$80.2 \\
$>$40 & 0 & -- & -- & -- & \multicolumn{1}{r@{\hspace{8pt}}}{--} & \multicolumn{2}{c}{--} \\
\addlinespace[3pt]
\multicolumn{8}{@{}l}{\textit{Branch depth}} \\
0 & 39 & 54.7 & 54.7 & 94.0 & $-$39.3 & $-$52.1 & $-$26.5 \\
1 & 45 & 40.7 & 40.7 & 90.4 & $-$49.6 & $-$60.7 & $-$37.8 \\
2--3 & 42 & 19.0 & 19.8 & 84.9 & $-$65.9 & $-$77.0 & $-$54.0 \\
4--6 & 58 & 10.3 & 11.5 & 82.2 & $-$71.8 & $-$78.7 & $-$64.4 \\
$>$6 & 56 & 1.8 & 3.0 & 86.3 & $-$84.5 & $-$89.9 & $-$78.6 \\
\bottomrule
\end{tabular}

\end{center}
\end{table}

\section{Scoring, estimands, and missing observations}
\label{app:statistics}

\textbf{Answer scoring.}
The harness extracts an \texttt{answer} field from a decodable JSON object.
Controlled circuits accept binary answers and controlled flowcharts accept
integer answers; public FlowGen uses nonempty string answers. Matching
normalizes Unicode with NFKC, case, whitespace, surrounding punctuation,
integer-like numeric strings, and simple option/choice labels. After this
normalization, exact equality is accepted. Comma-separated or
\texttt{||||}-separated lists are also compared as sorted normalized items.
No semantic judge or post-hoc synonym expansion is used. An accepted response
with no parseable answer is incorrect, not missing.

\textbf{Acquisition failure versus missing response.}
A schema-invalid representation is an observed pipeline failure: its
downstream question has score zero and no solver call. A rejected response,
such as a prohibited thinking marker, is instead unknown. The holdout
contains 11 rejected direct-vision outputs and no missing gold or learned
outcomes. The exposed recovery cohort contains three rejected direct-vision
outputs, two rejected gold responses, and no missing learned outcomes. These original observations are preserved without replacement.
No extracted string is selected using its downstream QA score.

For $N$ intended questions with $c$ observed correct answers and $m$ unknown
outcomes, the full-sample QA bound is $[c/N,(c+m)/N]$. For paired differences,
each unknown outcome is assigned its extreme binary value to obtain the
full-sample bound. Point estimates and bootstrap intervals use complete
pairs only. These bounds quantify missingness, not sampling uncertainty.
In particular, subtracting the displayed aggregate direct bound from a
full-cohort learned accuracy is not the same estimator as the 709-pair
comparison.

\textbf{Paired intervals.}
All questions and conditions belonging to the same chart move together in
10,000 bootstrap resamples, using seed 427. Per-bin QA differences are
ratios of summed paired scores to summed complete-pair counts within the
resampled charts. Difficulty slopes use equal-chart mean paired gaps
against the ordinal index of occupied eligible bins. Both slopes must have
the specified negative direction for the full difficulty hypothesis.
Intervals for the two slopes and the primary fidelity contrast use
Bonferroni 98.33\% limits; ordinary 95\% per-bin intervals remain descriptive.
The fidelity interval resamples fixed out-of-fold prediction losses,
conditional on the fitted folds, rather than refitting models inside every
bootstrap sample.

\begin{table}[t]
\caption{\textbf{Primary tests confirm a widening gap and improved relevant-topology
prediction.} QA slopes are pp per occupied ordinal bin, with equal chart
weight and fixed gold-floor eligibility. The Brier contrast uses the same
valid fixed-extraction cases. Estimates and paired chart-cluster CIs are
separate; group headings give the level of each interval. All entries are exports of the frozen analysis.}
\label{tab:inference}
\begin{center}
\small
\setlength{\tabcolsep}{4pt}
\begin{tabular}{@{}llr@{\hspace{8pt}[}r@{, }r@{]}}
\toprule
Outcome / contrast & Axis & \multicolumn{1}{c}{Estimate} & \multicolumn{2}{c}{CI} \\
\midrule
\multicolumn{5}{@{}l}{\textit{Primary tests (multiplicity-adjusted 98.33\% CI)}} \\
Fixed $-$ Gold QA & Node count & $-$23.1 & $-$30.7 & $-$15.6 \\
Fixed $-$ Gold QA & Branch depth & $-$11.2 & $-$14.8 & $-$7.6 \\
Relevant $-$ whole Brier & -- & $-$0.0259 & $-$0.0479 & $-$0.0047 \\
\addlinespace[3pt]
\multicolumn{5}{@{}l}{\textit{Secondary difficulty slopes (95\% CI)}} \\
Direct QA & Node count & $-$25.0 & $-$29.3 & $-$20.7 \\
Gold QA & Node count & $-$4.7 & $-$8.1 & $-$1.4 \\
Gold $-$ Direct QA & Node count & $+$20.3 & 15.1 & 25.5 \\
Direct QA & Branch depth & $-$11.5 & $-$13.9 & $-$9.0 \\
Gold QA & Branch depth & $-$2.4 & $-$4.0 & $-$0.8 \\
Gold $-$ Direct QA & Branch depth & $+$9.1 & 6.2 & 12.0 \\
\bottomrule
\end{tabular}

\end{center}
\end{table}

\begin{figure}[t]
\centering
\includegraphics[width=\linewidth]{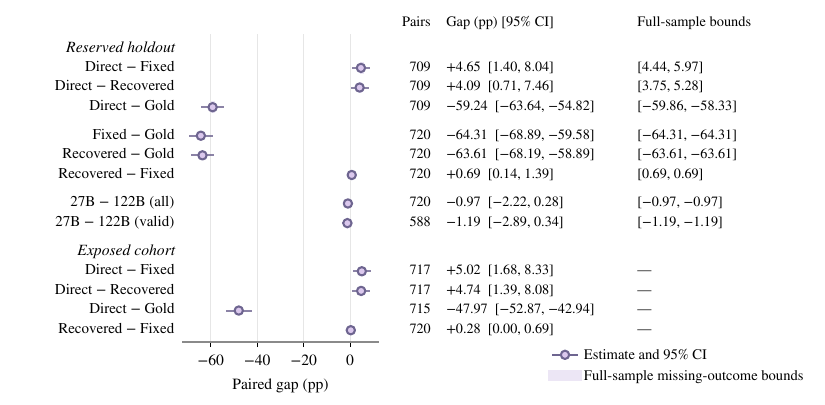}
\caption{\textbf{Paired holdout and exposed-cohort comparisons.} Points are paired gaps (pp) with 95\% chart-cluster intervals; gray bands are full-sample missing-outcome bounds, not confidence intervals. Holdout direct comparisons use 709 complete pairs (11 unknown observations); other holdout comparisons use 720 pairs, and the valid-input 27B comparison uses the identical 588 valid fixed-input questions. Exposed-cohort intervals are descriptive, at the aligned 8,192-token solver cap.}
\label{fig:paired}
\end{figure}

\textbf{Comparable accuracy and secondary trends.}
The predeclared margin is two absolute percentage points: a text condition
is comparable/noninferior to direct only when the lower paired 95\% bound
for text minus direct is strictly greater than $-0.02$. Non-significance of
a difference does not establish comparability. Gold passes this criterion;
fixed and recovered learned text do not. The secondary vision-difficulty
test compares equal-chart ordinal-bin slopes for direct and gold
(Table~\ref{tab:inference}). Gold declines more slowly along both axes,
supporting relative stability rather than a flat gold curve.

\subsection{Lenient answer-set scoring}
\label{app:lenient}

\textbf{Scoring robustness.}
The gold--direct--learned ordering survives partial-credit scoring.
We re-score the existing outputs of all five holdout arms without model
calls, changing neither the primary scores nor the confirmatory analysis.
For predecessor/successor questions, we compare sets of complete normalized
node labels, granting per-question set F1 or a binary match when Jaccard
similarity is at least 0.5. Labels use the original answer normalization;
we do not match isolated words within a label or introduce synonyms.
Relations and scalar shortest-path lengths retain exact scoring. Original
exact matches receive full credit. Invalid-input skips remain zero, and
the 11 rejected direct outputs remain missing.

\begin{figure}[t]
\centering
\includegraphics[width=\linewidth]{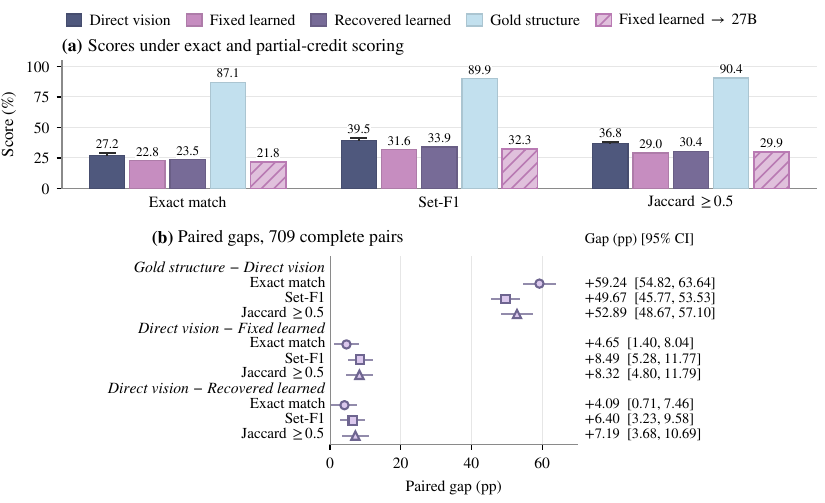}
\caption{\textbf{Lenient scoring preserves gold--direct--learned ordering.} (a) Five arms, 720 intended questions, 8,192-token solver cap. Set-F1 is macro-averaged; Jaccard uses a per-question threshold of 0.5. Direct whiskers denote missing-outcome bounds. (b) Paired gaps on 709 complete pairs with chart-cluster intervals. Both are post-hoc sensitivities.}
\label{fig:lenient}
\end{figure}

\section{Recovery policy and acquisition accounting}
\label{app:recovery}

\textbf{Bounded, validity-triggered acquisition.}
The primary arm always reports the original 1,024-cap extraction.
The secondary policy accepts that output if valid; otherwise it tries 2,048,
then 8,192. If the last output remains invalid, it applies a fixed
answer-free normalizer and, if needed, permits one 8,192-cap retry using the
same image plus the validation error and rejected transcription. The retry
instruction asks for a complete JSON object preserving visible primitives,
labels, and directions, not graph execution or an answer. The maximum is
four acquisition calls. Unresolved outputs remain invalid.

The normalizer checks schema validity first. It rejects
duplicate JSON keys and ambiguous identifiers. Where the structured schema
permits, an unspaced existing edge reference can justify an unambiguous
whitespace-only ID reconciliation; collisions are rejected. For controlled
circuits only, a redundant terminal \texttt{OUT} declaration can be removed
while retaining the output reference and every wire, provided there is no
outgoing edge from that terminal. These bounded format rules are not
semantic graph repair and are not inferred from reference answers. An
unsupported format or failed revalidation leaves the original invalid
representation unchanged.

\textbf{Observed public outcomes.}
On exposed public data, the original 218 valid charts become 240 valid
charts, with QA changing from 273/720 to 275/720. On the reserved holdout,
196 valid charts become 238, with QA changing from 164/720 to 169/720.
The two remaining invalid holdout acquisitions end in unterminated JSON;
they are retained as six intended QA failures. Recovery is therefore
successful on 99.2\% of public-holdout charts.

Selected successful holdout extractions emit a mean of 758.7 output tokens
(minimum 292, maximum 1,620; 238 charts). Cumulative acquisition, including
failed attempts, emits 1,136.4 tokens per intended chart on average
(minimum 292, maximum 19,456; 240 charts). The maximum is a sum over
attempts, not a single response exceeding its cap. There are 290 acquisition
attempts. Including every failed attempt and prompt gives a total
acquisition cost of 9,024.2 tokens per chart.

\textbf{Why this is not a uniform budget curve.}
Only invalid examples receive a larger output cap, and a valid but
semantically incorrect extraction is not retried. The policy therefore
answers whether bounded format recovery explains the public gap; it does
not estimate the accuracy of independently rerunning every chart at 8,192.
Actual token consumption, including every failed attempt, determines cost.

\section{Fidelity definitions and prediction analysis}
\label{app:fidelity}

\textbf{Alignment and whole-graph metrics.}
The offline adapter aligns emitted entity IDs to their own emitted labels,
never by fuzzy matching to the gold graph. Labels use NFKC normalization,
line-break removal, whitespace normalization, and case folding. Explicit
condition/label/text fields on a relation are retained; otherwise its
literal relation field is used, without synonym remapping. Unparsed
relation prose, ambiguous entity IDs, and unsupported relation structures
fail the adapter.

Directed topology exact is one if both the node set and directed-edge set
match the reference, ignoring edge labels. Directed-edge F1 compares sets of
ordered endpoint pairs. Labeled-edge F1 compares
$(\mathrm{source},\mathrm{label},\mathrm{target})$ triples.
Both-empty sets have F1 equal to one. Direction and label metrics are
reported separately rather than treating unlabeled topology as sufficient
for conditional questions.

\textbf{Question-relevant support.}
Support masks are deterministic functions of the question and released
source, used only offline. Incoming/outgoing-neighbor questions use all
edges incident to the queried target/source. Source edge relation questions
use the ordered pair in the question. Shortest-path questions use a
conservative BFS certificate: every outgoing edge of a source-reachable
node whose distance is less than the target distance, not merely one
successful shortest path. Unsupported or global questions retain the
whole graph. Wrong predicted incident neighbors count as false positives.
Unrelated isolated nodes are ignored by a local mask but remain errors for
whole-graph fidelity. These operational support masks may contain more
structure than a minimal sufficient subgraph.

\textbf{Matched prediction models.}
Relevant and whole versions use the same cases and five chart-held-out
folds. Each one-feature logistic model uses training-only scaling, L2
slope penalty 0.01, 1,000 optimization updates at learning rate 0.1, and
seed 427, with no tuning. Brier loss
$N^{-1}\sum_i(\hat p_i-y_i)^2$ is the primary metric; lower is better.
AUC is secondary. The primary population is the 588 fixed-learned
schema/adapter-valid observations. An all-intended sensitivity analysis
includes all 720 questions, with invalid representations assigned zero
fidelity predictors and QA zero. It is not substituted for the primary
matched-valid analysis.

\begin{figure}[t]
\centering
\includegraphics[width=\linewidth]{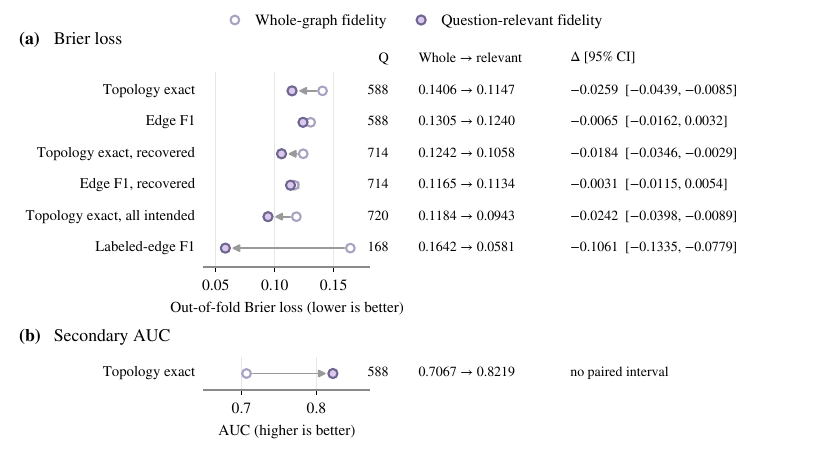}
\caption{\textbf{Matched fidelity prediction favors relevant topology.} (a) Whole (open) and relevant (filled) predictors use identical cases and folds; negative $\Delta$ favors relevance. Recovered and all-intended rows are sensitivities; labeled-edge F1 uses valid source-relation cases. (b) Secondary AUC, higher is better; no paired AUC interval was estimated.}
\label{fig:fidelity}
\end{figure}

Relevant labeled-edge F1 improves prediction on the valid source edge relation
subset, providing secondary label-sensitive evidence. The recovered-topology
Brier difference is $-0.0184$ (descriptive 95\% CI: $[-0.0346,-0.0029]$).
The fixed-extraction result remains the primary test; recovery is a
secondary sensitivity analysis.

\section{Complete exposed perturbation results}
\label{app:mechanism}

The exposed mechanism study uses a preselected 60-chart subset of the
240-chart exposed FlowGen cohort, evenly spaced in stored chart order.
Each relevant/irrelevant pair starts from the same source structure,
matching edit type and count. Feasible pairs are generated before solver
evaluation. Relevant edits
intentionally alter the original question's answer semantics; irrelevant
edits leave them unchanged. The original answer remains the scoring target.
The intervention measures sensitivity to the placement of answer-critical
corruption. Its effect size is conditional on the selected edits and does
not estimate how much of the natural acquisition gap these errors cause.

The reported FlowGen cells use Qwen3.5-4B, 27B, and 122B solvers with a
512-token output cap. They are separate from the final 8,192-cap holdout.
The number of eligible chart/question pairs varies across edit types and
edit counts. Intervals are paired chart-cluster 95\% bootstrap intervals,
10,000 resamples, seed 427; the exposed mechanism matrix is exploratory
and its many cell intervals are not multiplicity-adjusted.

\begin{figure}[t]
\centering
\includegraphics[width=\linewidth]{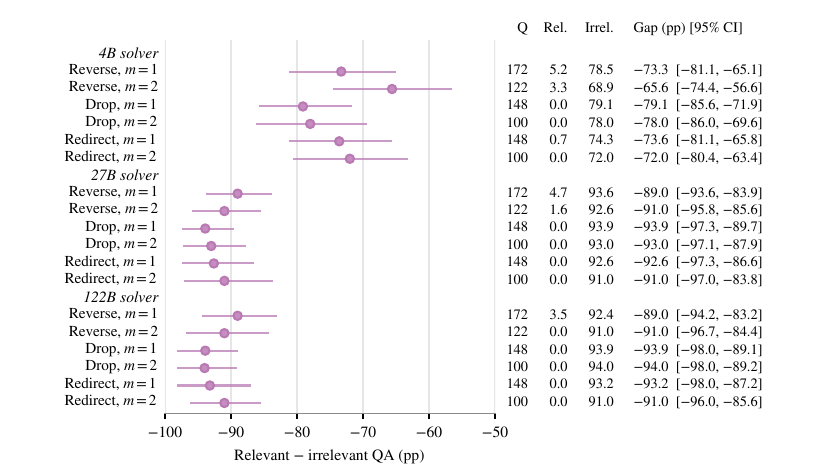}
\caption{\textbf{Relevant corruption reduces original-answer QA across matched edits.} Exposed source-input interventions at a 512-token solver cap. Gaps are relevant minus irrelevant with paired 95\% chart-cluster intervals; Q counts pairs and Rel./Irrel.\ are QA percentages. The 4B solver is lower even under irrelevant edits.}
\label{fig:perturbation}
\end{figure}

\section{Controlled post-confirmation diagnostic}
\label{app:controlled}

The combined controlled diagnostic contains 406 flowcharts and 135
circuits. The source-given controlled task and the public holdout are
different populations and question constructions. They are not pooled.
Controlled direct QA is already approximately 99\%; the original
node-count hypothesis did not confirm, and the valid subset did not
support estimating fidelity prediction. Recovery and the fixed
normalization were applied after confirmation and therefore serve as a
diagnosis of acquisition failures, not a new successful confirmation.

\begin{table}[t]
\caption{\textbf{Post-confirmation recovery approaches controlled direct-vision QA.}
QA (\%) at an 8,192-token solver cap; fixed columns retain original
extractions, recovery includes normalization. All 541 recovered charts are
valid, versus 111/406 flowcharts and 116/135 circuits at fixed 1,024.}
\label{tab:controlled}
\begin{center}
\small
\setlength{\tabcolsep}{4pt}
\begin{tabular}{@{}lrrrrrr@{}}
\toprule
 & & & \multicolumn{2}{c}{Fixed learned} & & \\
\cmidrule(lr){4-5}
Domain & Charts & Direct & 1,024 cap & 2,048 cap & Recovered & Gold \\
\midrule
Circuits & 135 & 99.3 & 85.2 & 89.6 & 99.3 & 94.1 \\
Flowcharts & 406 & 99.0 & 27.3 & 75.1 & 99.5 & 99.8 \\
\bottomrule
\end{tabular}

\end{center}
\end{table}

The recovered-minus-direct difference is $+0.49$ pp on flowcharts
(95\% CI $[-0.74,1.72]$) and zero on circuits
($[-2.22,2.22]$). Flowchart gold is 99.75\%; circuit gold is 94.07\%.
Gold denotes correct source input; answer correctness still depends on the solver.
This diagnostic and the public recovery result together distinguish
format/acquisition limitations in the controlled setting from a much
larger residual semantic extraction gap in the public setting.

\section{Token, call, and latency accounting}
\label{app:cost}

\begin{figure}[t]
\centering
\includegraphics[width=\linewidth]{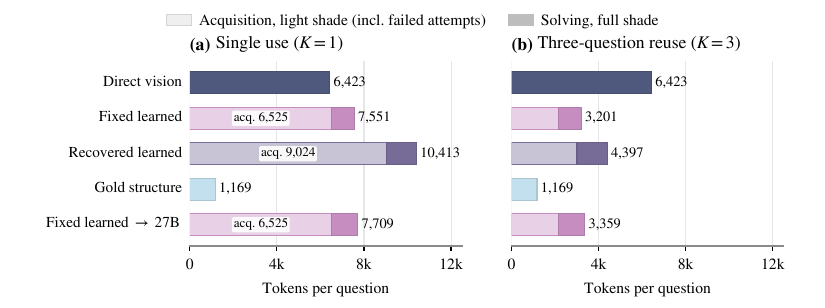}
\caption{\textbf{Actual native-token costs on the public holdout.} Means include all 720 intended questions and failed acquisition attempts. Light segments are acquisition (per chart in (a), amortized over the three questions in (b)); full-shade segments are solving; totals are printed at the bar ends. The identical fixed extraction is shared across the two solver arms.}
\label{fig:tokens}
\end{figure}

Table~\ref{tab:costmain} places accuracy and cost together; Fig.~\ref{fig:tokens} and Table~\ref{tab:latency} decompose that accounting into acquisition, solving, calls, and latency.

Let $A_g$ be all acquisition prompt and output tokens for chart $g$,
including failed attempts, and let $S_{gq}$ be solver tokens for question
$q$. Single-question logical cost is $A_g+S_{gq}$; for the three observed
questions on the same chart, average cost is
\[
 C_3(g)=\frac{A_g+\sum_{q=1}^{3}S_{gq}}{3}.
\]
The reported $K=1$ mean charges acquisition separately for each question;
$K=3$ amortizes the same acquired representation. Neither accounting
setting changes the generated answers. Invalid representations have no
solver call but retain their acquisition cost. Gold has zero modeled
acquisition cost only because the structure is supplied; its total is a
solver-only reference. Token totals are native model prompt plus
completion usage, not words, output caps, or a dollar-price comparison.

\begin{table}[t]
\caption{\textbf{Logical request counts and serial request latency.}
Entries are per intended question under single-use or three-question reuse.
Latencies reflect the execution hardware and serving conditions; they are
not a hardware-controlled solver-speed ranking. Cached reuse does not imply
new billed inference.}
\label{tab:latency}
\begin{center}
\small
\setlength{\tabcolsep}{4pt}
\begin{tabular}{@{}lrrrr@{}}
\toprule
 & \multicolumn{2}{c}{Calls per question} & \multicolumn{2}{c}{Seconds per question} \\
\cmidrule(lr){2-3}\cmidrule(l){4-5}
Input / solver & $K=1$ & $K=3$ & $K=1$ & $K=3$ \\
\midrule
Direct vision & 1.00 & 1.00 & 12.63 & 12.63 \\
Fixed learned & 1.82 & 1.15 & 19.35 & 10.71 \\
Recovered learned & 2.20 & 1.39 & 25.86 & 13.86 \\
Gold structure & 1.00 & 1.00 & 7.12 & 7.12 \\
\midrule
Fixed learned $\rightarrow$ 27B & 1.82 & 1.15 & 52.80 & 44.16 \\
\bottomrule
\end{tabular}

\end{center}
\end{table}

The holdout evaluation used 3,032 physical calls, including the 11 rejected
responses, and reused 588 identical solver responses without charging them
twice. Total actual prompt plus completion usage was 9,484,994 tokens.
The evaluation required 70.81 allocated GPU-hours over approximately
4.25 wall-clock hours; this allocation measure is distinct from active
device utilization.

\section{Question-family results and secondary scope}
\label{app:secondary}

\begin{figure}[t]
\centering
\includegraphics[width=\linewidth]{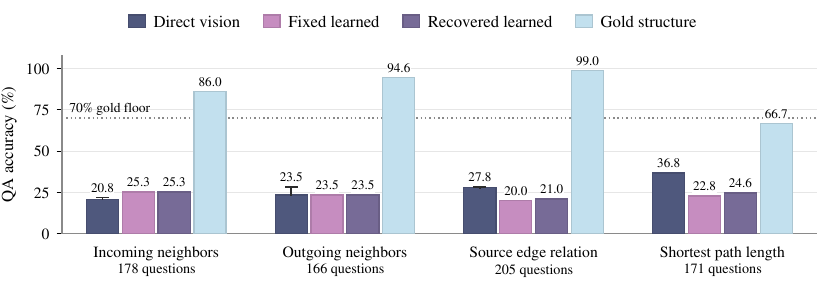}
\caption{\textbf{Shortest-path gold falls below the interpretive floor.} Exploratory QA by question family on intended holdout questions. Direct whiskers denote missing-outcome bounds; the dotted line marks the 70\% gold floor. These slices do not add confirmatory hypotheses.}
\label{fig:questiontypes}
\end{figure}

The public shortest-path family has gold QA 66.7\%, below the 70\% floor
used for structural-bin gap interpretation. This is reported rather than
removed; it does not invalidate the occupied node/depth bins, whose gold
QA remains above the floor. Source edge relation gold QA is 99.0\%, and the
label-sensitive fidelity comparison is reported separately.

\subsection{Source edge relation visibility and literal-label audit}
\label{app:visibility-audit}

The frozen relation-question generator selects a source triplet and asks
for its exact relation label; it does not check whether that string is
printed in the image or exclude generic source relations. We therefore
use \emph{source edge relation}, rather than \emph{printed edge relation},
for this question family. Of its 205 holdout questions, 124 on 115 charts
target \texttt{connectedTo}, and 18 target \texttt{partOf}. The remaining
63 relation questions have other source labels and have not undergone an
exhaustive image-visibility audit. These counts identify candidates by
reference label, not independently adjudicated visibility outcomes. An
adjudicated count would require item-by-item inspection of the original
images and source semantics, including unlabeled links and containment.

Original-image spot checks confirm unprinted source conventions. In easy
chart 105, the Cross\_selling Opportunities--Product Information Provision
link is unlabeled, as is the Solar Radiation Management Research--Climate
Education Programs link in easy chart 118; both reference answers are
\texttt{connectedTo}. In medium chart 129, Desalination is contained within
Water Meter Reading, rather than joined to it by a directed edge labeled
\texttt{partOf}. These examples show why source-triplet correctness is not
equivalent to recovering an explicitly printed relation. Source graph
topology can also encode containment conventions, so a visibility review
must examine graph semantics, not only remove generic answer strings.

On the 124 \texttt{connectedTo} questions, the recorded gold arm answers
123 correctly; direct, fixed learned, recovered learned, and the fixed
27B-solver arm each answer none correctly, with no missing responses on
this subset. The frozen aggregate and paired results are retained, not
rescored. Their interpretation includes source-annotation access as well
as acquisition and representation--solver effects. Any visibility-qualified
reanalysis would be a separately labeled post-hoc sensitivity, not a
replacement confirmatory result.

QA normalization applies Unicode NFKC, case normalization, whitespace and
specified answer-format handling, but does not equate internal hyphens and
underscores. Fidelity label normalization also retains this distinction;
node matching and directed-edge endpoints use normalized literal labels,
not fuzzy aliases. Thus \texttt{Third-Party Administrator Liaison} and
\texttt{Third\_Party Administrator Liaison} remain distinct under both
rules. No alias correction has been applied to the reported scores.

\subsection{QZhou as a secondary counterpoint}
\label{app:qzhou}

\begin{figure}[t]
\centering
\includegraphics[width=\linewidth]{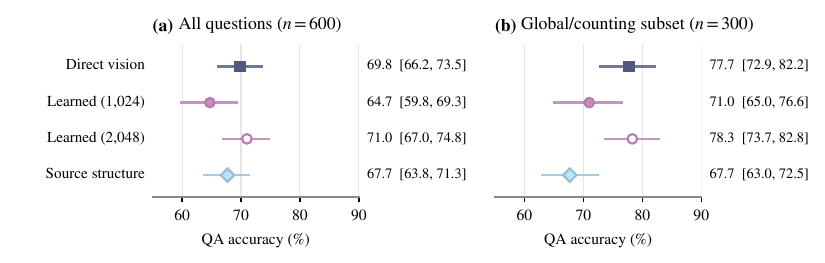}
\caption{\textbf{QZhou shows a different source-input pattern.} Exposed, secondary QA at a 512-token solver cap, counting invalid inputs as failures. Bars are paired-chart bootstrap 95\% intervals. Filled/open circles denote learned extraction at 1,024/2,048 tokens.}
\label{fig:qzhou}
\end{figure}

\textbf{Secondary comparison.}
Source-derived text is not uniformly better than direct vision on
QZhou. Fig.~\ref{fig:qzhou} reports the existing matched 200-chart cohort,
with three native questions per chart. Both extractor and solver are
Qwen3.5-122B-A10B. The extraction caps are 1,024 and 2,048; the common
solver cap is 512, unlike the final public FlowGen holdout. The
global/counting slice uses the previously assigned question taxonomy,
not a subset selected by correctness; it contains 300 questions on 187
charts. It includes global graph queries as well as counting and should
not be read as a pure arithmetic test.

The source-derived condition trails direct vision on the global/counting
slice, while a larger learned-extraction cap changes the ordering of the
learned arm. This is a solver/representation interaction at the recorded
cap, not evidence that source text is intrinsically insufficient or that
the model specifically fails at counting.

\textbf{Human checks and scope.}
The completed QZhou and stratified leakage audits are reported in
Appendix~\ref{app:human-audits}. The targeted QZhou sample does not establish
population source accuracy, and unreviewable leakage outputs remain unknown.
Query-conditioned outputs remain excluded from confirmation.
Historical learned-DOT small-solver results with an unresolved parser
version are not part of the confirmatory comparisons.

\section{Qualitative examples: which structure survives extraction?}
\label{app:qualitative}

Three post-analysis examples from the public holdout illustrate the
distinction between valid JSON, answer-critical structure, and a correct
answer. They are selected to explain contrasting outcomes, not to estimate
their frequency. All use the recorded question-blind Qwen3.5-122B
extraction and 8,192-token solver outputs. The figures are display-only
schematics of selected recorded source and extraction relations, not the
images used for evaluation. They resolve emitted entity IDs to their own
labels for readability. Original evaluation images, solver inputs, answers,
and reported scores are unchanged. Node shapes and positions in these
schematics carry no experimental information.

\subsection{A valid extraction can omit the edge needed for the answer}
\label{app:example-direction}

\textbf{Question.}
Which node labels are direct outgoing neighbors of ``Underwriting''?
The source answer is \emph{Legal Review of Claims} and \emph{Loss Prevention}.
The original prompt requires alphabetical labels separated by \texttt{||||}.

\begin{figure}[t]
\centering
\includegraphics[width=\linewidth]{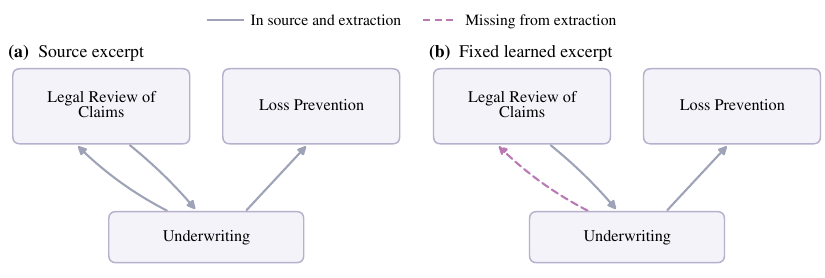}
\caption{\textbf{Valid extraction omits an answer-critical outgoing edge.}
FlowGen easy chart 63 (8 nodes, 8 edges). The source has opposite
directions between Underwriting and Legal Review of Claims; extraction
retains only the incoming direction. Display-only relation excerpts.}
\label{fig:qual-direction}
\end{figure}

\textbf{Structure comparison.}
Let U denote Underwriting, C Legal Review of Claims, and P Loss Prevention.
These abbreviations are used only in this display. The following entries
are selected relation records, not the full JSON. A dash means the
directed relation is absent; \texttt{connectedTo} is the source's generic
edge label, not necessarily printed text.

\begin{center}
\small
\begin{tabular}{@{}lll@{}}
\toprule
Directed edge & Source relation & Learned relation \\
\midrule
U $\rightarrow$ C & \texttt{educates} & --- \\
C $\rightarrow$ U & \texttt{advises} & \texttt{educates} \\
U $\rightarrow$ P & \texttt{connectedTo} & \texttt{Liaison} \\
\bottomrule
\end{tabular}
\end{center}

\textbf{Recorded answers.}
Direct vision and gold both return
\emph{Legal Review of Claims |||| Loss Prevention} (correct).
Fixed learned text returns only \emph{Loss Prevention} (incorrect).
Recovery reuses this already-valid representation and returns the same
answer. The solver's answer agrees with the outgoing neighbors in the
extracted structure, which is missing U $\rightarrow$ C.

\textbf{What this illustrates.}
The extraction used 422 output tokens, below its 1,024-token cap, and passed
the schema check. Both whole and relevant directed-edge F1 are 0.667;
relevant topology exact is zero. This is a semantic direction error that
validity-triggered recovery does not address, rather than a truncated JSON
object or a solver that ignores an available outgoing edge.

\clearpage
\subsection{An imperfect whole graph can preserve everything this question needs}
\label{app:example-relevant}

\textbf{Question.}
Which node labels are direct incoming neighbors of ``Digital Resource
Management''? The source answer is \emph{Library Website Management} and
\emph{Reference Desk Support}.

\begin{figure}[t]
\centering
\includegraphics[width=\linewidth]{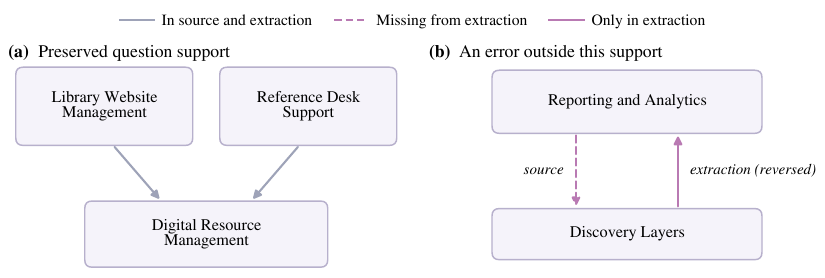}
\caption{\textbf{The two answer-critical incoming edges are preserved.}
FlowGen easy chart 290 (8 nodes, 8 edges). Extraction preserves both
incoming neighbors of Digital Resource Management despite reversing an
irrelevant edge. Display-only excerpts omit labels and other relations.}
\label{fig:qual-relevant}
\end{figure}

\textbf{Structure comparison.}
The learned JSON contains both relevant relations:
Library Website Management $\rightarrow$ Digital Resource Management,
and Reference Desk Support $\rightarrow$ Digital Resource Management.
It is nevertheless not an exact reconstruction. For example, outside the
question support, the source has Reporting and Analytics $\rightarrow$ Discovery
Layers (\texttt{digitizes}); the extraction reverses this edge. It also
reverses User Account Management $\rightarrow$ Reference Desk Support.
Neither edge enters the node asked about.

\begin{center}
\small
\begin{tabular}{@{}lrr@{}}
\toprule
Recorded fidelity & Whole graph & Question-relevant \\
\midrule
Directed-edge F1 & 0.667 & 1.000 \\
Directed topology exact & 0 & 1 \\
\bottomrule
\end{tabular}
\end{center}

\textbf{Recorded answers.}
Direct, fixed learned, recovered learned, and gold all return the two
correct labels, \emph{Library Website Management |||| Reference Desk
Support}. The fixed extraction is valid at 367 output tokens; recovery
therefore reuses it without another extraction call.

\textbf{What this illustrates.}
This example has the same whole-graph directed-edge F1 as
Example~\ref{app:example-direction}, but the question-relevant topology
is exact and learned QA succeeds. Whole-graph error alone does not tell
us whether the error affects the answer. The pair illustrates the
distinction tested by the aggregate matched-fidelity analysis; it is
not itself an additional test of predictive performance.

\clearpage
\subsection{Recovery can complete the JSON without repairing the relevant structure}
\label{app:example-recovery}

\textbf{Question.}
Which node labels are direct outgoing neighbors of ``Collaborative
Planning''? The source has five outgoing neighbors. The recovered
extraction preserves two, omits three, and adds a link to a container.

\begin{figure}[t]
\centering
\includegraphics[width=\linewidth]{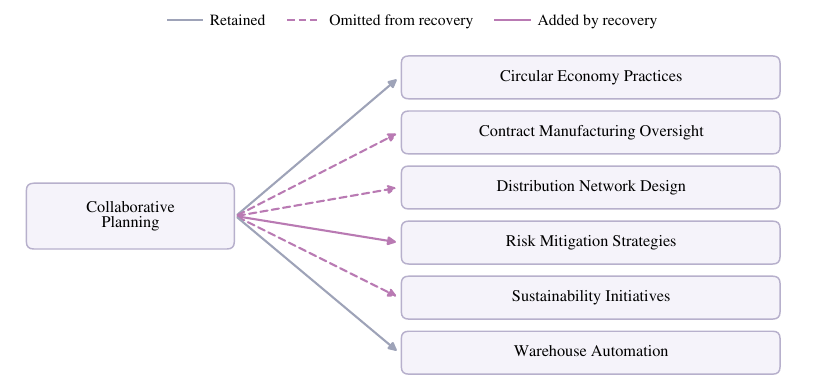}
\caption{\textbf{Valid JSON still confuses node and container destinations.}
FlowGen hard chart 108 (18 nodes, 38 edges). Gray edges are retained,
dashed pink edges omitted, and the solid pink edge incorrectly added to a
container. Display-only union of source and recovered outgoing neighbors.}
\label{fig:qual-recovery}
\end{figure}

\textbf{Recovery trace.}
At 1,024 output tokens the first JSON is truncated inside a relation record.
The 2,048-cap attempt completes in 1,176 output tokens and is schema-valid,
so no 8,192 extraction or bounded retry is needed. Full acquisition cost
is 19,102 prompt-plus-output tokens across both calls, including 2,200
output tokens; counting only the successful attempt would omit the first
call's cost. All solver calls use the common 8,192-token cap.

\textbf{Outgoing-neighbor comparison and answers.}
``Yes'' denotes membership in the target set. The gold answer matches the
source column; the recovered answer exactly matches its extracted
outgoing-neighbor set. The fixed arm is skipped for invalid input and
retains QA zero, rather than a fabricated solver answer.

\begin{center}
\small
\setlength{\tabcolsep}{9pt}
\begin{tabular}{@{}lccc@{}}
\toprule
Target label & Source/gold & Direct & Recovered \\
\midrule
Circular Economy Practices & Yes & --- & Yes \\
Contract Manufacturing Oversight & Yes & --- & --- \\
Distribution Network Design & Yes & --- & --- \\
Sustainability Initiatives & Yes & --- & --- \\
Warehouse Automation & Yes & Yes & Yes \\
Risk Mitigation Strategies & --- & Yes & Yes \\
\bottomrule
\end{tabular}
\end{center}

\textbf{What this illustrates.}
Gold is correct, whereas direct and recovered learned answers are incorrect.
Recovery raises schema validity but leaves relevant directed-edge F1 at
0.500 and relevant topology exact at zero (whole-graph edge F1: 0.738).
Here, the residual QA error is consistent with the missing and spurious
destinations in the completed structure, not with an incomplete JSON object.

\section{Solver-cap calibration and residual output hits}
\label{app:caps}

\textbf{Predeclared stopping decision.}
The exposed cap check treats a lower cap as saturated only if increasing it
changes QA by less than 3 pp and fewer than 5\% of outputs hit the lower cap.
The prespecified final check selected 4,096 if the rule was satisfied and
8,192 otherwise, with no further cap escalation. Both the
122B local-neighbor direct cell and the 27B source edge relation cell hit
4,096 on exactly 1/20 responses (5\%), which fails the strict
fewer-than-5\% criterion. The final common solver cap is therefore
8,192, without a claim that every question type is fully saturated.

For the 164 exposed path-gold cases, the 4,096 run has 126 known correct
answers and one missing response; the 8,192 run has 124 known correct and
two missing. On 162 complete pairs, the 8,192-minus-4,096 change is
$-1.23$ pp (95\% CI $[-6.79,4.32]$). Full-intended QA bounds are
76.83--77.44\% and 75.61--76.83\%, respectively. Cap hits are 8/163
and 6/162 accepted responses. The originally rejected response is treated
as missing, excluded from paired comparisons, and included in full-sample
bounds. Residual path errors after cap expansion are solver-side residuals;
the experiment does not isolate them as pure reasoning failures.

The local-neighbor direct-122B cell is 40\% at both final caps, and the
27B source edge relation cell is 35\% at both; each still has 1/20 cap hits at
8,192. The 512-cap experiments are budget-constrained exposed mechanism
or robustness results, not the confirmatory cap. The exposed difficulty
cohort was re-solved at the chosen cap using its existing extracted text;
no replacement extraction was generated for that cap-alignment comparison.

\begin{table}[t]
\caption{\textbf{Residual 8,192-cap hits and missing responses on the public
holdout.} A cap hit is a length finish or an output token count at the cap.
Rates use accepted solver responses. Invalid skips are observed QA failures;
missing responses remain unknown.}
\label{tab:cap}
\begin{center}
\small
\setlength{\tabcolsep}{4pt}
\begin{tabular}{@{}lrrrrr@{}}
\toprule
Input / solver & Accepted & Cap hits & Hit rate (\%) & Invalid skips & Missing \\
\midrule
Direct vision & 709 & 35 & 4.9 & 0 & 11 \\
Fixed learned & 588 & 11 & 1.9 & 132 & 0 \\
Recovered learned & 714 & 16 & 2.2 & 6 & 0 \\
Gold structure & 720 & 8 & 1.1 & 0 & 0 \\
\midrule
Fixed learned $\rightarrow$ 27B & 588 & 19 & 3.2 & 132 & 0 \\
\bottomrule
\end{tabular}

\end{center}
\end{table}

\begin{figure}[t]
\centering
\includegraphics[width=\linewidth]{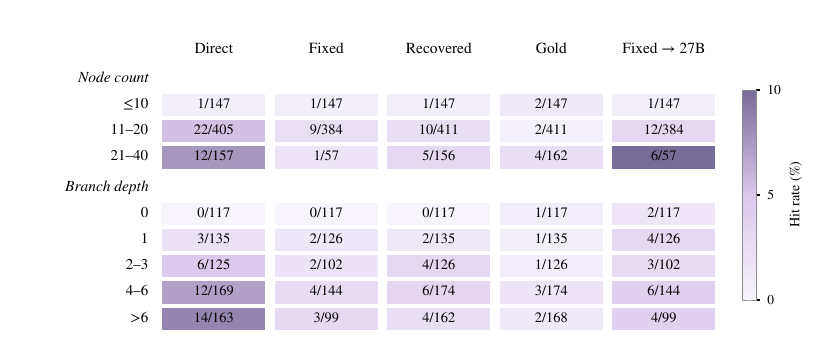}
\caption{\textbf{Cap-hit counts by prespecified structural bin.} Each cell is hits/accepted solver responses, not hits/intended questions; shading shows the hit rate. The empty $>40$ node bin has no observations.}
\label{fig:capbins}
\end{figure}

\section{Human audits of secondary evidence}
\label{app:human-audits}

\textbf{Samples and procedure.}
Three hired human reviewers independently reviewed each item in two
targeted audits. The QZhou packet contained 30 charts selected from a
model-flagged source--image discrepancy subset, with three native questions
per chart. The leakage packet contained 150 historical query-conditioned
textualizations: 100 ordinary outputs and 50 previously flagged or uncertain
outputs, stratified by domain, model, and protocol. Both samples are enriched
for suspected issues and do not estimate population rates.

Reviewers recorded verdicts, supporting evidence, confidence, and timestamps
without LLM-supplied judgments. For QZhou, they compared the image with the
released structure and separately assessed whether discrepancies could
affect each native answer. The leakage review provided the image, question,
verbatim textualization, and its historical contract, but withheld model
identity, prior flags, reference answers, QA scores, and gains. A contract
violation means that the text goes beyond permitted image-grounded evidence
to perform a prohibited downstream task, such as supplying the answer.
Malformed syntax alone does not constitute answer leakage. Independent
labels were saved before discussion and adjudication. The available review
summary does not specify whether disagreements were resolved by consensus,
majority vote, or a tie-breaker; we do not infer that procedure. Cases without
a resolved judgment remain unresolved or unreviewable. Denominators count
items, not judgments.

\begin{table}[t]
\caption{\textbf{Targeted human checks bound secondary interpretations.}
Counts are adjudicated item outcomes; three reviews per item do not multiply
sample size. Enriched sampling precludes population prevalence estimates.}
\label{tab:human-audit}
\begin{center}
\small
\begin{tabular}{@{}lr@{}}
\toprule
Adjudicated outcome & Items \\
\midrule
\multicolumn{2}{@{}l}{\textit{QZhou source--image consistency (30 charts)}} \\
\quad Match & 28 \\
\quad Cosmetic difference only & 1 \\
\quad Material mismatch & 0 \\
\quad Unresolved & 1 \\
\addlinespace[3pt]
\multicolumn{2}{@{}l}{\textit{QZhou answer relevance (90 questions)}} \\
\quad No observed answer-affecting discrepancy & 88 \\
\quad Answer-affecting discrepancy & 0 \\
\quad Unresolved & 2 \\
\addlinespace[3pt]
\multicolumn{2}{@{}l}{\textit{Query-conditioned leakage (150 outputs)}} \\
\quad No observed contract violation & 110 \\
\quad Contract violation & 5 \\
\quad Unreviewable & 35 \\
\bottomrule
\end{tabular}

\end{center}
\end{table}

\textbf{QZhou source--image consistency.}
Among 29 adjudicable charts, reviewers found no material mismatch: 28
matched and one differed cosmetically. One chart remained unresolved.
Of 90 question-level checks, 88 showed no observed answer-affecting
discrepancy and two were unresolved. Thus, observed source--image
disagreement does not account for the source-input errors on the
adjudicable audited cases. This finding neither verifies unaudited charts
nor estimates dataset-wide source accuracy. The QZhou comparison remains
secondary because it is exposed, uses a 512-token solver cap, and occupies
a single node-count bin.

\textbf{Query-conditioned contract review.}
Reviewers found five contract violations and 110 outputs with no observed
violation; 35 outputs were unreviewable and are not treated as clean. The
available review summary does not record item-level reasons for these 35
decisions, so we cannot attribute them to missing images, truncation, or any
other specific cause. They remain unknown.
The five violations among 115 reviewable outputs describe this enriched
sample, not a population leakage rate. All query-conditioned outputs remain
excluded from confirmation, including those with no observed violation.
The audit does not change the question-blind holdout or any frozen score.

\textbf{Agreement and reporting scope.}
Initial three-reviewer unanimity, before adjudication, was 25/30 (83.3\%)
for QZhou charts and 132/150 (88.0\%) for leakage items. Question-level
QZhou agreement and chance-adjusted agreement were not computed.
The audit summary does not report leakage outcomes separately for ordinary
and flagged strata, reasons for unreviewability, or whether the five
violations reached a solver; we make no claims about those quantities.
No rescoring or new inference was performed following the audit.

\end{document}